\documentclass[12pt]{amsart}

\usepackage{amscd}
\usepackage{marvosym}
\usepackage{amsmath}
\usepackage{amsgen}
\usepackage{amsbsy}
\usepackage{amsopn}
\usepackage{amsthm}
\usepackage{amscd}
\usepackage{amsfonts}
\usepackage{amssymb}
\usepackage{mathrsfs}
\usepackage{hyperref}
\usepackage{mathtools}
\usepackage{ifsym}
\usepackage{algpseudocode}
\usepackage{algorithm}
\usepackage{booktabs,array}
\usepackage[usenames,dvipsnames]{color}
		\definecolor{darkcyan}{rgb}{0., 0.65, 0.65}
        \newcommand{\cR}[1]{{\color{red} #1}}
        \newcommand{\cB}[1]{{\color{blue}{#1}}}

\usepackage[final]{graphicx}
\usepackage{todonotes}
\usepackage[marginratio=1:1,height=8.5in,width=6.5in,tmargin=1.25in]{geometry}
\usepackage[normalem]{ulem}
\usepackage{enumerate,enumitem}
\usepackage{tikz-cd}

\usepackage[utf8]{inputenc}
\usepackage{float}
\usepackage[caption = true]{subfig}
\usepackage{multirow}

\usepackage{comment}

\usepackage{latexsym}
\usepackage{MnSymbol}

\newcommand{\mc}{\mathcal}
\newcommand{\mbf}{\mathbf}

\newtheorem{Theorem}{Theorem}

\newtheorem{Proposition}[Theorem]{Proposition}

\newcommand{\op}{\operatorname}
\newcommand{\abs}[1]{\left| #1 \right|}
\newcommand{\norm}[1]{\left\| #1 \right\|}
\newcommand{\inner}[1]{\left\langle #1 \right\rangle}

\newcommand{\set}[1]{\left\{ #1 \right\} }
\newcommand{\tensor}{\otimes}

\DeclareMathOperator{\argmin}{argmin}
\DeclareMathOperator{\argmax}{argmax}

\newcommand{\R}{{\bf R}}

\newcommand{\eps}{\epsilon}

\newcommand{\la}{\lambda}

\newcommand{\cout}[1]{}

\renewcommand{\hat}{\widehat}

\def\be#1\ee{\begin{align}\begin{split} #1 \end{split}\end{align}}
\def\beq#1\eeq{\begin{align*}\begin{split} #1 \end{split}\end{align*}}

\title{DynACPD Embedding Algorithm for Prediction Tasks in Dynamic Networks}

\author[Connell]{Chris Connell}
\address[Chris Connell]{115 Rawles Hall, Indiana University, Bloomington, IN 47405 }
\email{connell@indiana.edu}

\author[Wang]{Yang Wang}
\address[Yang Wang]{Indiana University, Bloomington, IN 47405 }
\email{yw109@iu.edu}

\begin{document}

\begin{abstract}
Classical network embeddings create a low dimensional representation of the learned relationships between features across nodes. Such embeddings are important for tasks such as link prediction and node classification. In the current paper, we consider low dimensional embeddings of dynamic networks, that is a family of time varying networks where there exist both temporal and spatial link relationships between nodes. We present novel embedding methods for a dynamic network based on higher order tensor decompositions for tensorial representations of the dynamic network. In one sense, our embeddings are analogous to spectral embedding methods for static networks. We provide a rationale for our algorithms via a mathematical analysis of some potential reasons for their effectiveness. Finally, we demonstrate the power and efficiency of our approach by comparing our algorithms' performance on the link prediction task against an array of current baseline methods across three distinct real-world dynamic networks.
\end{abstract}

\maketitle

\section{Introduction}\label{sec:Intro}

Network embeddings provide a spatial representation of learned relationships between quantified feature data attached to the nodes and links of a network. While high dimensional embeddings may be important for sufficient resolution in some applications, most commonly the goal is to provide a low-dimensional representation of the learned feature relations, often for the purpose of efficient prediction or classification. The augmenting data may be commonly represented by weighted and possibly directed links.

However, many data sets carry a further temporal structure which is more naturally captured by a time-series of networks which we henceforth refer to as a {\em dynamic network}. (Dynamic networks also appear in the literature as ``temporal graphs [networks]'' or ``dynamical graphs [networks]''.) Examples of such data structures include internet/intranet networks, social networks, communication networks and scheduling protocols. In these examples, the nodes and links might represent websites and weblinks, people and relationships, site locations and connections or tasks and dependencies. Exploiting the temporal aspect of the data can be helpful in applications such as predicting unknown links ({\em link-prediction}), classifying communities of objects possessing a certain threshold of common linking ({\em node clustering}), and detecting outlier objects that do not belong to any clear community ({\em anomaly detection}). We will discuss each of these tasks in more detail in Section \ref{sec:tasks}.

As an approach to understanding static networks, spectral graph theory has been highly successful at capturing and quantifying various structural properties of a network at different scales. These methods consist of investigating the structure of networks by studying the eigenvalues and eigenspaces of linear operators associated to the network, such as its adjacency matrix or Laplacian matrix. Spectral invariants can distinguish certain information about local structure (e.g. connectivity and small cycles) from certain information about large scale structure in the network. Some of these invariants are physically motivated. For example, the concept of conductivity can be measured spectrally and used to deduce statements about connectivity. Other information besides connectivity is similarly encoded in the spectral data.

The spectral decomposition of an entire dynamic network, or at least partial decompositions for very large dynamic networks, can be performed. However this approach scales poorly for large networks and moreover aggregates the temporal aspects together with spatial ones. Thus their distinct temporal characteristics are lost. To remedy this we propose to apply higher order spectral methods to the dynamic networks which we encode as a 3-tensor of the individual Laplacian or adjacency matrices of each time-slice network. This becomes especially important in the case of directed network structures where the Laplacian matrix also encodes some of the information recovered by random walks on the network. There has been a fair amount of recent work relating tensor models of dynamic networks to various machine learning tasks (see e.g. \cite{BensonEtAl:15,AnandkumarEtAl:15,GleichEtAl:15,HuangEtAl:13,SunEtAl:06,XuEtAl:18,LuoEtAl:17,NguyenEtAl:18}).

\begin{figure}[htbp]
\centering
\includegraphics[width = 5in]{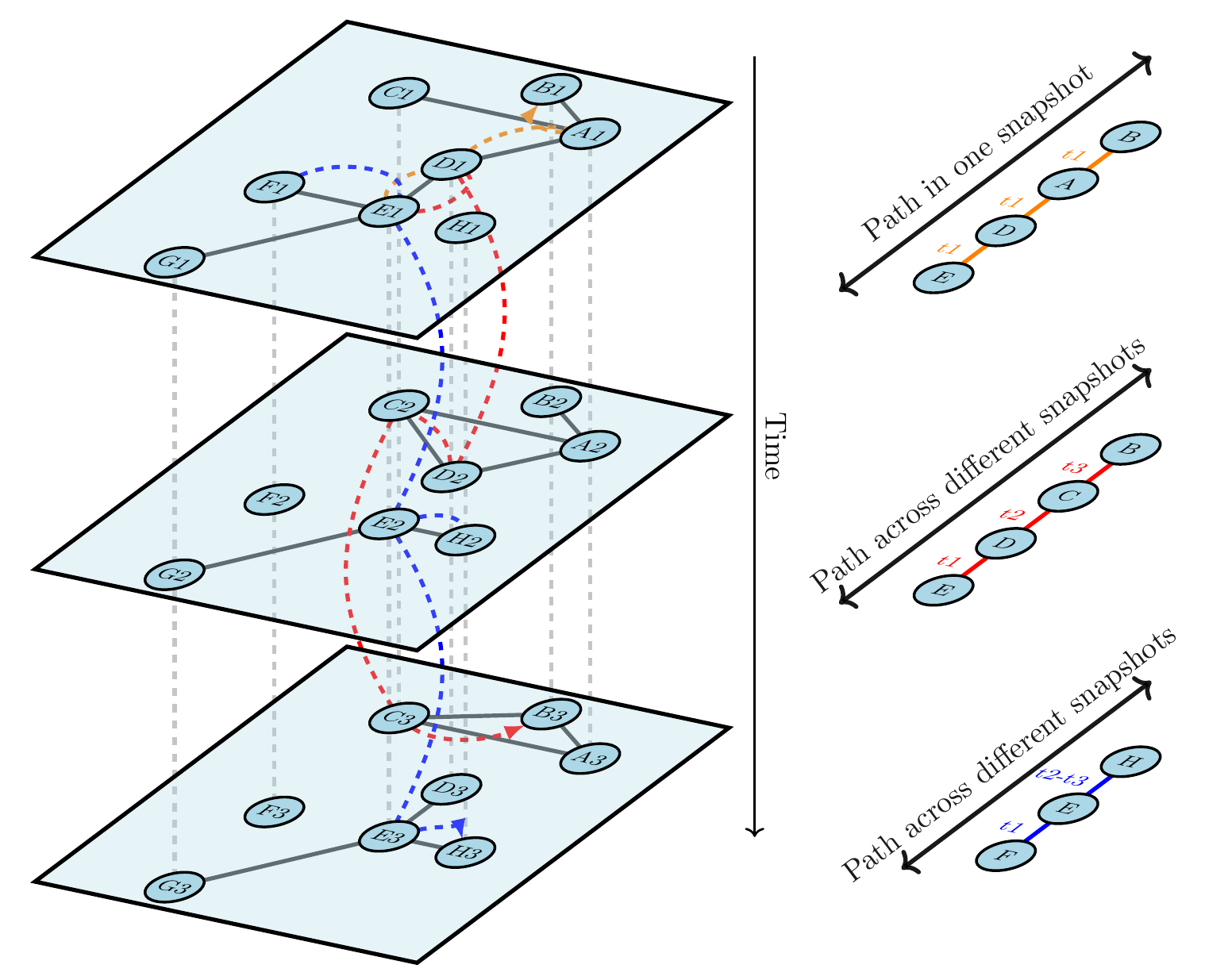}
\caption{Composite structure for the dynamic network.}
\label{fig:dynamic}
\end{figure}

We now specify the structure for a model of a dynamic network more precisely. Let $\mc{T}$ denote the indexed set of times used in our time series of (model) networks which we denote by $(G_t)_{t\in \mc{T}}$. We suppose that some nodes persist from one time to the next, so we assume they carry labels so that a node $v\in G_{t_i}$ is identified with a node in $G_{t_{i+1}}$ if they both carry the same label and that for any fixed time $t$ all the labels of $G_t$ are distinct. From this structure we may create a composite network $G$ which consists of the union of the (horizontal) time-slice subnetworks $\set{G_t}_{t\in \mc{T}}$ together with additional (vertical) links between nodes that carry the same node label (see Figure \ref{fig:dynamic}).

Without loss of generality, we may and do assume every time-slice network contains a node for every label in any time slice. We achieve this by adding in isolated nodes with the corresponding labels, optionally with an additional tag that indicates they were added in, to each time-slice network $G_t$ as necessary. Hence the number of nodes of each time slice is a constant number $\abs{V(G_t)}={n}$ being the totality of all labels. Setting ${\tau}=\abs{\mc{T}}$, where $\abs{\mc{T}}$ is the cardinality of $\mc{T}$, the composite network $G$ then has ${n}\times {\tau}$ nodes. Moreover, we will assume for simplicity that the times are self-indexed, namely $t_i=i$, as it is relatively simple to generalize formulas to the general case of unequally spaced times. (We will also assume by default that the most recent time is $\tau$.) Observe that, in this formulation, vertical links become redundant as they always exist between nodes. An alternate structure which we employ for unlabelled and unweighted networks, is again have all nodes in each network but to place a link from a node of $G_t$ to its copy in $G_{t+1}$ if and only if it exists at time $t$ in the dynamic network and persists to time $t+1$.

Next, we may reorganize time varying data with network relations as a 3-way tensor $Z$ of size ${n}\times {n}\times {\tau}$ whose slices transverse to the last dimension are the ${n}\times {n}$ adjacency matrices or Laplacian matrices encoding the network data at each corresponding time slice. (Recall that adjacency of nodes between time slices is automatic and therefore does not need to be additionally encoded.)

We wish to consider a more generlized dependency case where the connections are not automatically bidirectional and where the links are weighted with weights that perhaps reflect some preferences or strengths of connection. In this context our tensor $Z$ has components:
\[
Z(i,j,t)=\begin{cases} w_{ij} &\text{if node $i$ connects to node $j$ with weight $w_{ij}$  at time $t$} \\
0 & \text{otherwise}.\end{cases}
\]
Using this structure, our primary benchmark task will be the generalized link prediction problem which aims to predict the weights between nodes at times ${\tau}+1$, \dots, ${\tau}+k$ for some $k\geq 1$ given the entire structure $Z$ up to time ${\tau}$. 

While one can attempt to adapt two dimensional spectral methods to this framework, problems immediately arise. For example, spectral clustering methods attempt to find the fewest number of links to cut to isolate a cluster as a connected component, but here the complete set of connections between nodes in different time slices can confuse the operation of this procedure between the temporal dependencies and intra-network dependencies.

In order to extend the spectral methods to what now consists of an array of Laplacian or adjacency matrices we need a spectral method that applies to higher order tensors. Several such extensions exist (see the survey \cite{KoldaBader:09} and the references therein), though the mathematical theory behind such decompositions is neither as simple nor as complete as in the matrix case. For starters, the minimum number of summands, called the {\em rank} of the tensor, in a complete ``spectral'' (e.g. CP) decomposition of a tensor may not be constant on any dense open set of tensors of the same shape. Another problem is that even finding the rank of a tensor is NP-hard \cite{HillarLim:13}. Nevertheless, we are able to recover interpretable meaning from the corresponding tensor spectral methods which allows us to efficiently create low-dimensional embeddings which are very widely applicable to various machine learning tasks, including the three mentioned already. We also experimentally demonstrate the effectiveness of our embedding algorithms for the link prediction problem. The efficiency of our embedding algowithms rely on the fact that approximation schemes for tensor decompositions have been very efficiently implemented. We have developed modified versions of these standard algorithms to extend them to a context of weighted Frobenius norms and weighted tensors to both handle our more complicated context and provide the flexibility to potentially minimize various loss functions from our generalized learning tasks.

Attepmting to perform link prediction, node clustering, anomaly detection and other similar tasks directly on a dynamic network can suffer from dimensionality growth problems when the number of nodes is very large. Our approach to efficiently solving these tasks is based on an embedding of a combined representation of the dynamic network based on the history of the time slices. The embedding is optimized based on the (weighted and directed) generalized ``tensorial-spectral'' properties of the time slices to produce a geometric embedding of all nodes into a Euclidean space of a fixed and relatively small dimension $d$. This embedding is designed to reflect certain, depending on the precise embedding method, relational features between nodes. We aim for a certain proximity between nodes that are connected by a single link or multiple short paths so that the prediction tasks can be easily performed by a classifier based on a distance metric in $\R^d$.

In practice we treat the dimension $d$ as a hyperparameter to tune for a balance of task speed and accuracy, and we have found that even for dynamic networks with more than $1,500$ nodes an embedding dimension of $d\leq 32$ still provides excellent resolution together with outstanding time performance (see Section \ref{sec:experiments}). These embeddings gives a geometrically proportional representation of the expected relations within nodes of the network at the current time $t=t_\tau$ (the point at which predictions are requested).

Before describing our tensor embedding algorithms we will recall in the next section some of the important static network embedding methods based on the spectral theory of graphs as this theory will serve both as guiding principle and the backbone for our tensor embedding methods.

\section{Spectral Graph Embeddings}\label{sec:Spectral}

We recall that the (ordinary) Laplacian of a graph or network $G=(\mc{V},\mc{E})$ is given by the matrix $L=D-A$ (normalized version $\mc{L}=I-D^{-\frac12}AD^{-\frac12}$) where $D$ is the diagonal matrix with diagonal element $D_{vv}=d_v$, the degree of node $v$, $I$ is the identity matrix and $A$ is the adjacency matrix (here we make the exception that $(D^{-1})_{vv}=0$ if $d_v=0$, corresponding to an isolated node). The eigenvalues of $L$ are always nonnegative since $L$ factors as $L=M^TM$ for the corresponding incidence matrix $M$ of the network. Observe also that in the case the network is regular of degree $d$, the $\la$-eigenvectors of $\mc{L}$ are the $d(1-\la)$-eigenvectors of $A$. 

For a static network, we recall that the eigenvectors $\set{f_1,\dots,f_{n}}$ of $\mc{L}$, form a basis for $\ell^2(G)$. Writing the eigenvalues, counted with multiplicity from least to greatest, for $\mc{L}$ as $\hat{\la}_1\leq \hat{\la}_2\leq \cdots\leq  \hat{\la}_n$ it is known that these lie in the interval $[0,2]$ and the number which vanish coincides with the number of connected components of the network (See \cite{Chung:96}). The Raleigh principle allows us to express the eigenvalues as,
\[
\begin{aligned} \hat{\lambda}_{k} &=\inf _{f} \sup _{g \in P_{k-1}} \frac{\sum_{u \sim v}(f(u)-f(v))^{2}}{\sum_{v}(f(v)-g(v))^{2} d_{v}} \\ &=\inf _{f \perp P_{k-1}} \frac{\sum_{u \sim v}(f(u)-f(v))^{2}}{\sum_{v} f(v)^{2} d_{v}} \end{aligned}
\]
where $P_{k-1}$ is the subspace spanned by all eigenfunctions $f_i$ for $i \leq k-1$. If $\hat{\la}_s=\hat{\la}_{s+1}=\dots=\hat{\la}_{t}=\hat{\la}$ then the $f_j$ are no longer unique up to scale for $s\leq j\leq t$, however, we may assume we have chosen the $\hat{\la}$-eigenfunctions so that they form a fixed orthonormal family, which then span the $\hat{\la}$ eigenspace.

These formula give us an interpretation of the $f_k$ as resonant functions, or modes, on $G$ and the $\hat{\la}_k$ are the corresponding strength of those resonances or modes on $G$.

In particular, knowing the strongest modes tells us about the neighbors. We may interpret the eigenfunctions (eigenvectors of $\mc{L}$) as an analogous Fourier basis for real valued functions on $G$.  The $0$-eigenfunctions are the harmonic functions which are constant on the components of finite networks. This represents the fact that the total mass transferred into a node by the Laplacian from neighboring nodes equals the total mass transferred out of that node to neighboring nodes on any finite network.

\subsection{Adjacency minimized embeddings}\label{subsec:AdjacencyEmbed}
We observe that for eigenfunctions of small eigenvalues the values at nodes must be similar when most of their neighbors are similar since the net flow into the two distinguished nodes from the common neighbors must nearly cancel the outflow. Viewed from the opposite perspective, the eigenfunctions for small eigenvalues distinguish highly connected clusters. This suggests that using the coordinates of the top $d$-eigenfunctions as an embedding of the network into $\R^d$, where $d<<\abs{\mc{V}}={n}$ is a reasonable choice of solution of initial embedding if our goal is to group connected vertices together. Similar embeddings and their applications have been investigated from a number of directions (see e.g. \cite{PachevWebb:17,CaiEtAl:07,CaiEtAl:18,HeEtAl:04}). To understand more precisely why this idea works, consider a possibly generalized adjacency matrix $A$ whose entry $A_{vw}$  represents a proximity weight between nodes $v$ and $w$ (the larger the weight, the closer the nodes should be in their underlying feature space). Now consider the case $d=1$, so we wish to embed the network into $\R$ with relative distances preserved as much as possible. One approach is to formulate this as the following weighted $L^2$-minimization problem,
\begin{equation}\label{eq:Eigmin}
y^{*}=\argmin \sum_{v,w\in\mc{V}}\left(y_{v}-y_{w}\right)^2A_{vw}=\argmin 2 y^{T} L y,
\end{equation}
where we normalize the scaling of $y$ with the additional constraint that $y^{T} D y=1$, as this emphasizes that the larger the value of $D_{v v}$, the more important $y_{v}$ is. (Here we used that $D_{vv}=\sum_{v\neq w} A_{vw}$.) Observe that this problem is equivalent to the following optimization,
\[
y^{*}=\argmin_{y^{T} D y=1} y^{T} L y=\argmin \frac{y^{T} L y}{y^{T} D y}=\argmax \frac{y^{T} A y}{y^{T} D y}.
\]
In other words $y^*$ is just the largest eigenvector for $D^{-1}A$ (which need not be symmetric). Note that after transforming this eigenvector by the fixed linear change of variables $y^*= D^{-\frac12}z^*$, then 
\[
D^{-\frac12}AD^{-\frac12}z^*=D^{\frac12}D^{-1}Ay^*=D^{\frac12}\la y^*=\la z^*,
\]
where $\la$ is the eigenvalue associated to $y^*$.
Thus the linearly transformed $z^*$ is the eigenvector of largest eigenvalue for the symmetric matrix $D^{-\frac12}AD^{-\frac12}$, or equivalently the eigenvector of smallest eigenvalue, corresponding to eigenvalue $1-\la$, for $\mc{L}$. The desired embedding is the coordinate embedding $i\mapsto {z^*}_i\in \R$.

For the case of higher dimensional embeddings ($d>1$) we inductively perform the minimization in \eqref{eq:Eigmin} over the orthocomplement of the span of the previous minimizers. That is, given that we have found minimizers $y_1^*,\dots,y_i^*$ we choose $y_{i+1}$ to be the minimizer of \eqref{eq:Eigmin}, except that we restrict the $\argmin$ to the subspace $y_{i+1}\in P_i=\left(\op{span}\set{y_1^*,\dots,y_i^*}\right)^\perp$. 

Again, after transforming by the linear map $D^{\frac12}$, we obtain vectors $z_1^*,z_2^*, \dots,z_d^*$ which are the first $d$ smallest eigenvectors of $\mc{L}$. The corresponding embedding is $i\mapsto ({z_1^*}_i,{z_2^*}_i,\dots,{z_d^*}_i)$. In particular, this embedding is linearly equivalent to the adjacency minimized embedding whereby we use the coodinates of the eigenvectors corresponding to the $d$ largest eigenvalues. This adjacency matrix formulation is often preferable to the Laplacian formulation for computational reasons simply because it is easier to easier to find and work with the largest part of the spectrum than the smallest part. Lastly, the overall scaling of the embedding is not important in principle, so we may use unit eigenvectors for the embedding or any other overall scaling factor that is convenient for applications.

Considering third order, or higher order, averaging operators could potentially improve the embedding as this will take into account adjacent nodes to the adjacent nodes as well. Investigations in this direction for different contexts include \cite{AbuElHaijaEtAl:19,RossiEtAl:18,AbuElHaijaEtAl:18,RoblesKellyHancock:07}. However, some of these approaches involve an additional time penalties due the additional arithmetic complexity involved.

\subsection{Generalizations to Weighted Directed Graphs}

For many applications we wish to allow networks where nodes of our network have various properties associated to them. (Consider, for example, a social network where various characteristics of participants are listed with each individual who is represented by a single node.) There are various approaches to representing the effects of such additional property data. One common approach is to ``one-hot'' each node into a vector based on characteristics, and then measure node distances based on some utility norm applied to the resulting vectors. This process results in a weighted network, and if the relationship between nodes is not assumed to be reflexive then the result is a directed weighted network.

If we consider such a directed weighted network with link weights $\set{w_{ij}}$ on links connecting nodes $i\to j$ then the adjacency matrix is simply $A=[w_{ij}]$. The weighted degree matrix is the diagonal with entries $D_{i i}=\sum_{k} w_{ik}$. 

In this case there are at least four different competing definitions for the directed Laplacian, each useful for different purposes. The simplest two are to use the same formulae as in the directed case but with either in or out degrees:
\[
L=D_{in}-A \quad\text{or}\quad L=D_{out}-A
\]
and
\[
\mc{L}=I-D_{in}^{-\frac12}A D_{in}^{-\frac12} \quad\text{or}\quad  \mc{L}=I-D_{out}^{-\frac12}A D_{out}^{-\frac12}
\]
One problem that arises with these choices is that the lack of symmetry in $A$ which is now reflected in $L$ and $\mc{L}$ means that there are no longer only real eigenvalues. However they still admit a singular value decomposition, so we can obtain the singular values and pairs of eigenrepresentatives instead of individual eigenfunctions. 

We will indicate why a more useful symmetric formulation of the Laplacian will be practical for the network embedding and node prediction problem (see \cite{Chung:05}). We define the symmetric Laplacian and the normalized symmetric Laplacians (respectively) as 
\[
\hat{L}=\Phi-\frac12 (\Phi P+ P^T\Phi)\quad\text{and}\quad \hat{\mc{L}}=\Phi^{-\frac12} \hat{L} \Phi^{-\frac12}=I-\frac12\left( \Phi^{\frac12} P\Phi^{-\frac12}+\Phi^{-\frac12} P^T\Phi^{\frac12}\right),
\]
where $P$ is the probability transition matrix of $G$ given in terms of the weights $w_{ij}$ by $P_{ij}=\frac{w_{ij}}{\sum_{k}w_{ik}}$ and $\Phi$ is the diagonal matrix with diagonal elements $\phi_i$ where $\phi=(\phi_i)$ is the unique left-eigenvector of $P$ with all positive entries normalized by the requirement that $\sum_i \phi_i=\sum_{j,k} w_{jk}$. This eigenvector necessarily has eigenvalue $1$, i.e. $\phi.P=\phi$. While $\phi$ does not have a closed form description which is independent of ${n}$, it nevertheless can be found to high accuracy in sub-cubic time via the Perron-Frobenius iteration method.

In the undirected (symmetric weights) case $\phi=(\phi_i)=(d_i)=(\sum_k w_{ik})$ and so $\Phi=D, P=D^{-1}A$ and $P^T=AD^{-1}$. This yields $A=\frac12{\Phi P+P^T\Phi}$ and therefore we have $\hat{L}=L$ and  $\hat{\mc{L}}=\mc{L}$, i.e. these definitions reduce to the previously defined ones. Note that in this case, we may explictly express
\[
\hat{\mc{L}}_{ij}=\mc{L}_{ij}=\delta_{ij}-\frac{w_{ij}}{\sqrt{\sum_{k}w_{ik}}\sqrt{\sum_{k}w_{jk}}}=\delta_{ij}-\frac{\frac12(w_{ij}+w_{ji})}{\sqrt{\sum_{k}w_{ik}}\sqrt{\sum_{k}w_{jk}}},
\]
where $\delta_{ij}$ is the Dirac Delta function.

Given these considerations, to extend the spectral embeddings described in the previous subsection to the undirected weighted case, we simply use $\frac12(\Phi P+P^T\Phi)$ in place of $A$ and $\Phi$ in place of $D$. The minimization problem in \eqref{eq:Eigmin} is otherwise the same.

\subsection{Generalized Adjacency Matrices}\label{subsec:GenAdj}

Consider the generalized symmetrized weighted adjacency matrix $A:=\frac12(\Phi P+P^T\Phi)$ corresponding to the symmetric Laplacian for the possibly directed network $G$ described in the previous subsection. Starting from this ``adjacency matrix'' $A$, we may further modify it in order to capture certain distinct node connection information in our embeddings.

The (modified) Katz metric counts all paths between two nodes, discount longerer paths by a decaying exponential in their length. Define $\op{P}_{i  j}^{\ell}$ to be the set of all paths of length $\ell$ from node $i$ to node $j$. Then given a weight $0<\omega <1$ we may formally define,
\[
\operatorname{Katz}(i,j)=\sum_{\ell=1}^{\infty} \omega^{\ell-1}\left|\operatorname{P}_{i  j}^{\ell}\right|
\]
The corresponding network kernel can be formally expressed in a closed form as,
\[
A_\omega=\frac1\omega\left((I-\omega A)^{-1}-I\right)=\sum_{\ell=1}^{\infty}\omega^{\ell-1}A^{\ell},
\]
where $A$ is the usual adjacency matrix of $G$. Provided $\omega$ is strictly smaller than the inverse of the spectral radius $\rho(A)$ of $A$, then the above formal sums converge. For the unweighted and undirected case, the sharp upper bound for the spectral radius given in \cite{HongEtAl:01} allows us to choose any value of $\omega$ satisfying,
\[
0\leq \omega < \frac{2}{\delta-1+\sqrt{(\delta+1)^{2}+4(2 m-\delta n)}}
\]
where $\delta$ is the minimum degree of a node and $m$ is the number of links. This bound can be easily expoited to obtain a  na\"ive (but not-so-sharp) upper bound for the allowable $\omega$ parameters in the general directed and weighted case given in terms of largest and smallest weights.

Notice that $\lim_{\omega\to 0}A_\omega=A$. Whenever $A$ is nonnegative or symmetric, then $A_\omega$ is as well. Moreover, we can view $A_\omega$ as  a weighted adjacency matrix for a virtual network $G^{(\omega)}$ with $G^{(0)}=G$. 

The main purpose of this generalization is to include multistep adjacencies, exponentially dampened by the number of steps in the path, when considering link prediction and similar problems. The same embedding algorithm used on these matrices will then reflect this variant of proximality information in the embedding. We will use the $A_\omega$ matrices for the time-slice networks in our DynA(O)CPD algorithms described below, viewing $\omega\geq 0$ as a tunable parameter.

\subsection{Resistance Embeddings}\label{sec:resist}

There is an important variant of the adjacency minimized embeddings described in Section \ref{subsec:AdjacencyEmbed} above. This embedding takes into account not only the eigenvectors of the smallest eigenvalues of $L$ but also a specific weight involving the eigenvalues. This embedding separates points according to idealized resistance between nodes when the network is viewed as an electrical circuit. For this reason the resulting embedding is usually called the {\em Resistance Embedding}. 

As indicated earlier, for an ordinary (unweighted, undirected) network $G=(V,E)$ with $n=|V|$ nodes, the kernel of its adjacency matrix $A$ has dimension equal to the number, $c$, of connected components of $G$.  For the ordinary Laplacian $L$ we may compute its unique $n\times n$ pseudo-inverse (Moore-Penrose inverse) $L^{\dagger}$. Since $L$ is a positive semi-definite symmetric matrix this will simply be $L^{\dagger}=O\Sigma O^T$ where $\Sigma$ is the diagonal matrix with diagonal values $\left(\frac{1}{\la_1},\dots,\frac{1}{\la_{n-c}}\right)$ where $0<\la_1\leq \la_2 \leq \dots\leq\la_{n-c}$ is the nondecreasing sequence of nonzero eigenvalues of $L$ and $O$ is the $n\times n-c$ rectangular matrix whose $n-c$ column vectors $\set{v_i}$ are an independent set of $\la_i$-eigenvectors for $L$. (While $O$ is not unique, any two choices produce the same $L^\dagger$, which is unique.)

Following \cite{PachevWebb:17}, the {\em commute time} is defined to be
\[
C_{x, y}=|E|\left(L_{x x}^{\dagger}+L_{y y}^{\dagger}-2 L_{x, y}^{\dagger}\right),
\]
where the quantity $r_{x, y}=\left(L_{x, x}^{\dagger}+L_{y y}^{\dagger}-2 L_{x, y}^{\dagger}\right)$ is known as the {\em effective resistance} of the network.  

Indeed, $r_{x, y}$ is precisely the electrical resistance between nodes $x$ and $y$ where the network represents an electrical circuit and each link a resistor with resistance given by its inverse weight in the adjacency matrix (see \cite{DoyleSnell:84} and the references therein for related results and further discussion). More recently, it has been shown in \cite{GhoshEtAl:08} that effective resistance is, in fact, a metric distance function on the network.

Let $S=\sum_{i=1}^d \frac{1}{\la_i} v_i \tensor v_i$ be the $n\times n$ rank $d$ approximant to $L^{\dagger}$. By the standard theory of elliptic operators, $S$ minimizes the $L^2$-operator norm $\norm{S'-L^{\dagger}}_{op}$ among all matrices $S'$ of rank at most $d$, and moreover $\norm{S-L^{\dagger}}_{op}= \frac{1}{\la_{d+1}}$ for $d<n-c$. The formula for the resistance embedding map $f_R:\mc{V}\to \R^d$ is then given by 
\[
f_R(j)=(\frac{{v_1}_j}{\sqrt{\la_1}},\dots,\frac{{v_d}_j}{\sqrt{\la_d}}).
\] 
The following effective reformulation of Proposition 3.1 of \cite{PachevWebb:17} explains the relationship between these. 

\begin{Proposition}\label{prop:resist}
Let $d$ be a positive integer and let $G=(V, E)$ be a connected, undirected network. Then for $S$ and $f_R: V \rightarrow R^{d}$ as above, we have that for all $i, j \in V$, 
\[S_{i  i}+S_{j j}-2 S_{i j}=\|f_R(i)-f_R(j)\|_{2}^{2} .\]
\end{Proposition}

We observe that the proof given in \cite{PachevWebb:17} works equally well in the more general disconnected case as well, provided that we define everything as we have done.

Lastly, we wish to address the time complexity of the resistance embedding. Just as the $k$ largest eigenvalues and eigenvectors of $L$ can be approximated efficiently and inductively via Perron-Frobenius/Power methods, the $k$ smallest eigenvalues and eigenvectors can also be found efficiently using the fact that $L$ is positive semi-definite. Namely, if $\la$ is the largest eigenvalue of $L$ then $\la+\mu$ is the smallest eigenvalue of $L$ where $\mu$ is the largest eigenvalue of $L-\la I$. Inverting these small eigenvalues gives us the top $k$ eigenvalues, together with their corresponding eigenvectors, for $L^\dagger$ that we need for the resistance embedding. This approach avoids the large time expense, currently greater than $O(n^{2.3})$, for computing the entirety of $L^\dagger$ from $L$. 

\section{Tensor Decompositions}\label{sec:CPD}
Before introducing our embedding algorithms and their functional explanation and justification, it will be necessary to review some aspects of general tensor decompositions.

\subsection{Polyadic Decompositions}
Given a standard Euclidean $k$-tensor, $Z\in \R^{n_1}\times\cdots \R^{n_k}$, for large enough $r$ we may decompose $Z$ as
\begin{align}\label{eq:decomp}
Z =\sum_{i=1}^{r} \lambda_{i} \mathbf{a}_1^{i} \tensor \cdots \tensor \mathbf{a}_k^{i},
\end{align}
for a set of generalized scalars $\la_i\in \R$, called {\em modes}, and unit vectors $a_j^i\in \R^{n_j}$ called {\em mode vectors}. Such a decomposition is called a {\em polyadic decomposition}. We define the {\em (tensor) rank} of $Z$, denoted $r(Z)$ to be the smallest number such that there is a decomposition of the form \eqref{eq:decomp}. While the rank of matrices is always at most $2$, it can be difficult to compute the tensor rank of $k$-tensors for $k>2$. As mentioned in the introduction, the number $r(Z)$ is NP-hard to compute (\cite{HillarLim:13}) with respect to the entry parameters as shape dimensions increase. Moreover, for a fixed shape $(n_1,\dots,n_k)$, $r(Z)$ may vary over $Z$ even taking on distinct values on disjoint open sets (see \cite{Strassen:83,CatalisanoEtAl:02}). If shape dimensions are ordered such that $n_1\geq n_2\geq\cdots\geq n_k$ then an upper bound for the tensor rank is $r_E(n_1,\dots,n_k)=\prod_{i=2}^k n_i$. However, in most cases, the maximum rank is expected to be much smaller, namely at most $2\frac{\prod_{i=1}^k}{\sum_{i=1}^k (n_i-1)}$ and half of that on a Zariski dense open set (\cite{AboEtAl:09}).

Whenever the number of terms in a decomposition \eqref{eq:decomp} is minimal (i.e. the tensor rank), it is called a {\em canonical polyadic decomposition (CPD)}. Other names for the CPD are the Canonical Decomposition (CANDECOMP) and the Parallel Factors (PARAFAC) decomposition. (There are two common choices regarding the modes $\la_i$: if we insist that the $\la_i\geq 0$, then this generalizes the notion of a singular value decomposition, and otherwise this decomposition generalizes the eigen-decomposition for matrices that are diagonalizable over $\R$.)

We say a decomposition \eqref{eq:decomp} is {\em unique} if it is unique up to permutation of the summands among all decompositions with the same $r$. The {\em identifiable} tensors are those for which the CPD is unique. Apart from some special shape-rank cases, and provided the rank is not too large, tensors are generically identifiable, meaning there is a (Zariski) open dense set of identifiable tensors of that shape and rank. For an identifiable tensor, the modes $\la_i$ are unique and the singular directions are unique provided the $\la_i$ are distinct. (Unlike the case of matrix decompositions, for $k$-tensors with $k>2$ the decomposition is often still unique when some modes $\la_i$ coincide but depends on the dimensions and choice of $Z$.) As is well known, matrices are never identifiable and hence extra conditions on the $a_1^i$ and $a_2^i$ are typically imposed.

Certain $k$-tensors $Z$ of a given rank can even be expressed as the limit of a sequence of lower rank tensors. (In this case some of the $\la_i$ in the sequence become unbounded). This poses a challenge for estimates and numerical algorithms alike.

The lack of uniqueness of a CPD for a nonidentifiable $k$-tensor $Z$ can be treated by placing enough additional assumptions on the decomposition, provided these are not so constraining that they fail to exist. Finding natural conditions can be subtle. For example, one might hope that if the matrices $A_\ell=[ a^1_\ell \dots a^r_\ell]$ consisting of column vectors of the decomposition \eqref{eq:decomp} have full rank for each $\ell=1,\dots,k$ then $Z$ is identifiable. However, as mentioned earlier this fails for almost all matrices, but does work generically for many (but not all) rank and shape combinations provided $k\geq 3$ (\cite{KoldaBader:09}). The next section provides a natural normalization which does (essentially) address this issue.

\subsection{Orthogonal Canonical Polyadic Decompositions}

A decomposition of the form \eqref{eq:decomp} with minimal $r$ such that for one index $\ell$ the vectors $\set{a^1_\ell,\dots,a^r_\ell}$ form an orthonormal set in $\R^{n_\ell}$ is is sometimes called an {\em $\ell$-orthogonal CP decomposition}. Moreover, in this case the modes $\la_i$ become bounded even for any approximating sequence. The downsides to imposing such a constraint are that the minimal $r$ needed for an exact decomposition may be necessarily larger than the tensor rank under this assumption, and the running time of existing numerical approximation algorithms for the constrained decomposition is longer. (However, as we demonstrate experimentally in Section \ref{sec:experiments},  neither downside is of significant concern in practice, at least for the tensor sizes that arise in our examples.)

Similarly, one may insist for any subset of the component indices $S\subset \set{1,\dots,k}$ that the corresponding $\set{a^1_j,\dots,a^r_j}$ are orthonormal in $\R^{n_j}$ for all $j\in S$. The case when $S=\set{1,\dots,k}$ is called a {\em completely orthogonal CP decomposition (COCPD)}. If we only insist that in \eqref{eq:decomp} either $\inner{a^i_\ell,a^j_\ell}=0$ or $a^i_\ell=\pm a^j_\ell$ for every distinct pair $i,j\in {1,\dots,r}$ and all $\ell\in \set{1,\dots,k}$ for minimal $r$ then we say that the decomposition is a {\em strongly orthogonal CP decomposition (SOCPD)}. A SOCPD always exists while a COCPD may not exist, even for $k=3$ (\cite[Corollary 3.9]{Kolda:01}). However, SOCPDs need not be unique. Indeed, let $a,b\in \R^m$ be two orthonormal vectors, then the 3-tensor 
\[
Z_1=4 a\tensor b\tensor b + 3  b\tensor b\tensor b +  a\tensor a\tensor a
\]
is presented as an SOCPD.  It is easy to check that the tensor $Z_1$ also admits the distinct SOCPD,
\[
Z_1=5 (\frac45 a+\frac35 b)\tensor b\tensor b + \frac45 (\frac45 a+\frac35 b)\tensor a\tensor a +  \frac35 (\frac35 a-\frac45 b)\tensor a\tensor a.
\]
If we order the $\la_i$ by $\la_1\geq\la_2\geq \cdots \geq \la_r$ then when computing a SOCPD we may further insist that we choose the largest possible $\la_1$, then the largest possible $\la_2$ and so forth. Let us call such a choice of ``maximal'' SOCPD simply an {\em orthogonal CP decompostion (OCPD)}. An OCPD is unique if and only if the $\la_i$ are all distinct (see the discussion after Corollary 3.9 of \cite{Kolda:01}).

We define the {\em orthogonal tensor rank} to be the minimal $r$ in \eqref{eq:decomp} for an OCPD. (Note that here we insist that $r\geq \min{n_1,\dots,n_k}$.) Observe that the orthogonal tensor rank is the same as the analogously defined strongly orthogonal tensor rank. 

The OCPD will be the tensor decomposition we employ in our DynAOCPD algorithm described in Section \ref{sec:algorithms}. While we primarily use the CPD and OCPD decompositions in this paper, we remark that there are closely related decompositions such as the {\em Tucker decomposition} whereby we write,
\[
Z=\mathcal{K} \times_{1} U^{(1)} \times_{2} U^{(2)} \times_{3} \cdots \times_{k} U^{(k)}
\]
where $\mc{K}$ is an $r_1\times r_2\times\cdots\times r_k$ matrix called the {\em core} matrix and the $U^{(i)}$ are $n_i\times r_i$ factor matrices and $\times_{i}$ indicates contraction of $U^{(i)}$ with $\mc{K}$ along the $i$th dimension. The {\em Higher Order Singular Value Decomposition (HOSVD)} is an important special cases of this decomposition where we take the core to be the same shape as $Z$ and require that the $U^{(i)}$ are orthogonal $n_i\times n_i$ matrices. Note that the HOSVD can be expressed in the form \eqref{eq:decomp} where $r=n_1\times n_2\times\dots\times n_k$ and the $\la_i$ are the entries of $\mc{K}$ and $a^i_j$ is the appropriately indexed row of $U^{j}$. However, the HOSVD differs from the OCPD in having both a greater number of terms and in that the $a^i_j$ must be rows of (square) orthogonal matrices. Nevertheless, the CPD and OCPD can also be interpreted as Tucker decompositions with approriate choices for the shape and entries of the core $\mc{K}$. Further discussion and numerical comparison of algorithms for these decompositions can be found in \cite{BatselierWong:17} and \cite{RabanserEtAl:17}.

\subsection{Approximations via Generalized Frobenius Norm}
We recall that the Frobenius Norm is the norm on $\R^{n_1}\times\cdots \R^{n_k}$  given by,
\[
\norm{\sum_{i_1,\dots,i_k=1}^{n_1,\dots,n_k} c_{i_1,\dots,a_k}e_{i_1}\tensor \cdots\tensor e_{i_k}}=\sum_{i_1,\dots,i_k=1}^{n_1,\dots,n_k} \abs{c_{i_1,\dots,a_k}}^2
\]
which is induced from the following inner product (called the Frobenius inner product) on basic tensors:
\[
\inner{\mathbf{v}_1 \tensor \cdots \tensor \mathbf{v}_k,\mathbf{w}_1 \tensor \cdots \tensor \mathbf{w}_k}=\prod_{i=1}^k \inner{v_i,w_i}_i,
\]
where $\inner{\cdot,\cdot}_i$ is the standard inner product on $\R^{n_i}$.
Note that for all flavors of orthogonal CP decompositions described in the previous section the individual summands $\mathbf{a}_1^{i} \tensor \cdots \tensor \mathbf{a}_k^{i}$ are Frobenius-orthonormal to $\mathbf{a}_1^{j} \tensor \cdots \tensor \mathbf{a}_k^{j}$ for all $j\neq i$. Indeed, in each case at least one mode vector in the $i$-summand is orthogonal to its corresponding mode vector in the $j$-summand (for the SOCPD case observe that the summands must be linearly independent).

For our application to the dynamic adjacency tensor $Z$ we will only need $k=3$ with a common first two dimensions so we will simplify notation by setting $n_1=n_2={n}$ and $n_3={\tau}$, and we wish to find a ``best'' approximation, 
\[
Z \approx \sum_{i=1}^{d} \lambda_{i} \mathbf{a}_{i} \tensor \mathbf{b}_{i} \tensor \mathbf{c}_i.
\]
for $d<r(Z)$ where $r(Z)$ is either the minimal rank or the orthogonal rank of $Z$ depending on the desired decomposition.

The most common way to find this approximation, and the one we will use, is called the Alternating Least Squares (ALS) method. We refer the reader to \cite{SidiropoulosEtAl:17,KoldaBader:09} and \cite{RabanserEtAl:17} for details. As indicated above, we will insist on the $\mathbf{a}_i$,$\mathbf{b}_i$ and $\mathbf{c}_i$ being unit vectors so that ALS approximation becomes generically unique. One advantage of the standard ALS algorithm is that it can be fairly easily modified to find an orthogonal CPD approximation to Z as well (see Section \ref{sec:experiments} and code for details of our implementation of this modified algorithm).

The ALS algorithm relies on the Frobenius norm as its optimization metric. We  use a modified Frobenius norm to find a best choice of $B={\mathbf{b}_1 \dots \mathbf{b}_r}$ matrix for the embedding where we may selectively emphasize more recent data over past (potentially stale) data in the dynamic network. 

Our modified Frobenius norm will simply use a weighted inner product in the $t$-domain, the inner product on $\R^{\tau}$ becomes
\[
\inner{u,v}=\sum_{i,j=1}^{\tau} w_{ij} u_i, v_j,  
\]

where the matrix $[w_{i j}]$ is positive definite. In practice we will use diagonal weights that favor more recent time slices: $w_{\tau}>w_{{\tau}-1}>\dots>w_1>0$.

Note that with respect to the standard $L^2$-Frobenius norm, we can achieve the same effect of the weighted norm by changing the tensor being normed by the transformation $a_{i j k}\to a_{i j k}\sqrt{w_{i j k}}$.

Recall that the spectral decomposition of a symmetric network Laplacian matrix (or adjacency matrix) provides an $L^2$-orthogonal basis of eigenvectors viewed as functions on the nodes of the network. Similarly, we will call the last two tensor components, $\mathbf{b}_{i} \tensor \mathbf{c}_{i}$, of the $i$-th summand in a general CP decomposition the $i$-th {\em mode function}. Observe that mode functions can be identified with functions on the complete dynamic network (or rather our augmented version of it). Under some additional mild assumptions on the generalized adjacency or Laplacian matrices used as slices, the complete set of mode functions for any of the various (full rank) CP decompositions provides a modified Frobenius-$L^2$ basis for the space of Frobenius-$L^2$ functions on the nodes of the dynamic network. (Since we drop the first vector of the decomposition summands, some mode functions may be dependent on linear combinations of others so one iteratively drops the dependent ones with smallest mode until a basis is achieved.) Moreover, the (full rank) OCPD corresponds to an orthogonal basis (akin to a two dimensional discrete Fourier basis) for this space of functions, ordered by strength of the modes.  This functional point of view will be a helpful guide in understanding these tensor decompositions from a generalized spectral vantage point.

\section{Dynamical Spectral Embeddings}\label{sec:algorithms}

For the case of dynamic networks, recall by our prior considerations that we may assume without loss of generality our time-slice networks each have $n$ nodes and our corresponding tensor $Z$ has shape $n\times n\times \tau$ with each (horizontal) time slice consisting of the $n\times n$ generalized adjacency matrix for the time-slice network as described above.

Our goal is to use the modified (orthogonal or free) CP decompositions described above to find embeddings which take into account both temporal coupling and static time adjacencies. Later in this section (\ref{sec:Just}), we show that the mode vectors corresponding to the largest modes behave similarly to the matrix case. Namely they will have similar coordinate values on highly correlated common neighbors. (Note that some coordinates without many common neighbors but that happen to have contributions that are similar may agree for a given mode function, but will not agree at all mode scales.)

The coupling network $G_C$ has node set corresponding to all nodes at any time and all links between adjacent time slices, but none of the links within a time slice. Hence the degrees of nodes in $G_C$ are always $0,1$ or $2$. Our underlying assumption is that most nodes will have adjacent neighbors as only a relatively small fraction of nodes are first seen or last seen in a given time-slice network. Therefore the average node degree of $G_C$ is close to, but less than, $2$.

Note that link weights in $G_C$ are not captured in the $Z$ tensor representation. However, when a node does not have degree $2$ in $G_C$, then either it, its predecessor or its successor has degree 0 in its corresponding time-slice network, and hence has no intra-slice link weights. Consequently, the mode functions will detect the orphaned node, and so we are not fundamentally losing information about the coupling in the dynamic network, unless we wish to consider the case of fully weighted coupling links.

The first stages of our algorithm is a pre-processing step and a post-processing step on the data to account for temporal degredation of data. These are explained in the next two subsections.

\subsection{Pre-Conditioning of Data via Weighted Linear Recurrence}
For the data pre-processing step, we build a weighted flow on the data from more recent time slices to older ones. As we want to minimize the time cost of this step we utilize a simple linear recurrence to achieve this flow based on a given a vector of temporal weights $(w_t)_{t\in \mc{T}}$ on the time slices with $w_t\in [0,1]$. For these weights, we use the normalization $\ell_\infty(w_t)=1$. (Recall also that we assumed that the most recent time corresponds to $t=\tau$.)  We flow network information back to create a new vector of networks $(G_t^{\mathrm new})$  from the original vector of networks $(G_t^{\mathrm old})$ by the following linear recurrence (computed inductively):
\[
G_\tau^{\mathrm new}= w_\tau G_\tau^{\mathrm old},
\]
and
\[
G_t^{\mathrm new}=  w_t G_t^{\mathrm old} + (1-w_t)G_{t+1}^{\mathrm new},
\]
where for a weighted network $G$ and a scalar $\alpha$, the network $\alpha G $ is simply the network all of whose link weights are multiplied by $\alpha$. For most applications we would set $w_\tau=1$. Moreover if all of the weights are $w_t=1$, then $G_t^{\mathrm new}=G_t^{\mathrm old}$ for all $t\in \mc{T}$. 

The general case of weights can be considered to be a smoothing process of flowing more recent information backwards via a time dependent averaging operator. One important application of this step when performing link prediction is to suppress the effects of localized large variance in link weights caused by high frequency link changes. Moreover, this scheme is flexible enough to allow us to exclude individual time slices completely by setting the corresponding weight $w_t=0$. One of the main points of incorporating weights on the data in this particular way is that the overall scale of the weights on the links is not affected by this process: if a link weight $w_{ij}$ occurs in every $G_t^{\mathrm old}$ for $t\geq t_0$, then $G_{t_0}^{\mathrm new}$ will have the same link weight $w_{ij}$ as well, independent of the choice of network weights $(w_t)$. 

Typical examples of the preprocessing weights used include decreasing exponentials, $w_t=e^{-\alpha \abs{\tau-t}}$, and normalized Gaussian weights, $w_t=e^{-\frac{(\tau-t)^2}{2\sigma^2}}$. 

\subsection{Post-Conditioning via Spectral Temporal Weighting} 
In addition, it is convenient to assign (or learn) temporal weights to the mode decomposition as a post-processing. We may think of these weights $\hat{w}=(\hat{w}_t)_{t\in\mc{T}}$, possibly distinct from the $(w_t)_{t\in\mc{T}}$, as weights on the $\mathbf{c}_i$ vectors. We normalize these weights by $\ell_\infty(\hat{w}_t)=\max_{t\in\mc{T}} \hat{w}_t=1.$ In the (O)CP decompositions, for a fixed $t\in \tau$ the coefficient ${\mathbf{c}_i}_t$ of $\mathbf{c}_i$ provides a weighting of the modes $\la_i$ which may reorder the relative importance of the mode vectors in the reconstruction of the slice $Z_t$. In particular, ${\mathbf{c}_i}_t=\frac{1}{\sqrt{{\tau}}}$ for all $i$ and $t$ if and only if all of the $Z_t$ coincide and hence the $\mathbf{b}_i$ can be chosen as the common eigenvectors. 

Since it may be that a large $\la_i$ has small ${\mathbf{c}_i}_t$ components for recent values of $t$, and we consider only $d$ many components, we will want to order the decomposition components by size of $\sigma_i=\la_i \inner{\hat{w},\mathbf{c}_i}$. Whether we take the largest or the smallest depends on the choice of $Z_t$ as we will see. The point is to always select the most important modes with respect to the given choice of temporal weighting.

\subsection{The DynA(O)CPD and DynL(O)CPD Embeddings}
If $V=\set{v_1,\dots,v_{n}}$ represents the entirety of nodes of the weighted networks $G_t$ then we form the embedding over $i\in \set{1,\dots,{n}}$, 
\[
v_i\mapsto (\sigma_1 {\mathbf{b}_{1}}_i,\sigma_2{\mathbf{b}_{2}}_i,\dots,\sigma_d {\mathbf{b}_{d}}_i)
\]
where ${\mathbf{b}_j}_i$ is the $i$-th component of the $j$-th tensor decomposition vector $\mathbf{b}_j$ and $\sigma_i=\la_i^\frac12 \inner{\hat{w},\mathbf{c}_i}$ is the $i$-th weighted mode for the CP (respectively, OCP) decomposition of the tensor $Z$ with $Z_t=A_t$, the corresponding adjacency matrices of the slices. The square root on the $\la_i$ appears because we intend to use the $L^2$ norm for distance comparisons. (Note this is the same reason the $-\frac12$ power appears on the $L$-eigenvalues of the resistance embedding.)

Using the corresponding coordinates of the top $d$ $\mathbf{b}$-NCP-eigenvectors thus gives an embedding which reflects similar static adjacencies on average, but modified to acknowledge strong temporal neighborhood similarity. How this information arises and how it is extracted will be explored in the next section.

Analogously, the DynLCPD (resp. DynLOCPD) embedding uses the tensor $Z_t=L_t$. However, in this case we use the embedding,
\[
v_i\mapsto \left(\frac{\inner{\hat{w},\mathbf{c}_{r(Z)}}}{\la_{r(Z)}}{(\mathbf{b}_{r(Z)})}_i,\frac{\inner{\hat{w},\mathbf{c}_{r(Z)-1}}}{\la_{r(Z)-1}}{(\mathbf{b}_{r(Z)-1})}_i,\dots,\frac{\inner{\hat{w},\mathbf{c}_{r(Z)-d+1}}}{\la_{r(Z)-d+1}} {(\mathbf{b}_{r(Z)-d+1})}_i \right).
\]
In other words, we use the smallest nonzero (temporally weighted)  modes and correpsonding decomposition vectors, and we weight the components of the eigenvectors with the inverse of the modes in parallel with the Resistance embedding for the single matrix case. As will be pointed out in the next section the DynL(O)CPD embeddings, at least in principle, capture the same network informations as the DynA(O)CPD embeddings. As they are also significantly more time consuming to compute we will principally focus on the DynA(O)CPD embeddings for experimental comparisons in Section \ref{sec:experiments}.

\subsection{Mathematical Justification of The DynACPD and DynAOCPD Embeddings}\label{sec:Just}

In this section, we provide a theoretical basis for why our DynA(O)CPD (dynamic adjacency canonical polyadic decomposition) embedding algorithm is effective, and generally outperforms the baseline embedding methods as demonstrated by the experiments of the next section.

Let ${Z}_t$ be a matrix associated to the network of the $t$-th time-slice. For instance, we could take ${Z}_t$ to be one of $A_t$,$L_t$ or $\mc{L}_t$. We can write this $Q_t$ in terms of that of the final time slice, namely ${Z}_t={Z}_{\tau}+\Delta_t$ for some matrix $\Delta_t$. Suppose moreover that,
\[
\frac{\norm{\Delta_t}_2}{\norm{{Z}_{\tau}}_2}\leq C \epsilon,
\]
where the constant $C$ depends on the structure of the network and especially on its degree vector (local connectivity) and $\eps>0$ is small. When $Q_t$ is one of the three matrix families mentioned above, then the assumption that only $\eps n$ nodes change in the unweighted network case will cause the assumption to be satisfied. In the weighted case, where the weights of $\Delta_t$ are at most $\eps$ times those of $\norm{{Z}_{\tau}}_2$, then we still have the same bound, but for perhaps a larger constant which also depends on the maximum of the weights. 

If $\lambda^t_i$ and $e_i^t$ are the eigenvalues and orthonormal choice of eigenvectors of ${Z}_t$, then we have by estimates (e.g. see \cite{IpsenNadler:09,Thompson:76}) for symmetric matrices that
\[
\la^t_i-\la^{\tau}_i=-\la_i^{\tau}(e_i^{\tau})^T\Delta_te_i^{\tau} +O(\eps^2\norm{{Z}}_2^2)
\] 
and 
\[
e_i^t-e_i^{\tau}=-e_i^{\tau}\left(\frac12 (e_i^{\tau})^T\Delta_t e_i^{\tau}\right)-\la_i^{\tau}\sum_{j\neq i} \frac{(e_j^{\tau})^T\Delta_te_j^{\tau}}{\la_i^{\tau}-\la_j^{\tau}}e_j^{\tau}+O(\eps^2\norm{{Z}}_2^2).
\]
Moreover,  the Bauer-Fike Theorem gives $\abs{\la_i^t-\la_i^{\tau}}\leq \norm{\Delta_t}_2$. (These formulas have nonsingular versions when the eigenspaces are not one dimensional.)

Hence the large eigenvalues and corresponding eigenvectors of ${Z}_t$ are controlled by those of ${Z}_{\tau}$. While the eigenvectors suffer greater sensitivity, they do not change much on eigenspaces where  the matrices $\Delta_t$ are nearly $0$ on the eigenspaces of nearby eigenvalues.

Writing the (O)CPD of our combined ${Z}=({Z}_t)_{t\in\mc{T}}$ tensor as,
\[
{Z} =\sum_{i=1}^{r} \lambda_{i} \mathbf{a}_{i} \tensor \mathbf{b}_i \tensor \mathbf{c}_i,
\]
we can use the above small changes assumption to estimate each of these. 
We first note that in the case when the tensor has constant slices ${Z}_t={Z}_{\tau}$, i.e. $\eps=0$, then $r={n}$ and the $\la_i=\sqrt{\tau}\la_i^{\tau}$, $\mbf{a_i}=\mbf{b_i}=e_i^{\tau}$ and  $\mbf{c_i}=\frac{1}{\sqrt{\tau}}(1,1,1,\dots,1)$. In the small changes case, 
we have $\abs{\lambda_i-\sqrt{\tau}\la_i^{\tau}}\leq \sqrt{\tau}\sup_t \norm{\Delta_t}_2$, and provided the $\la_i^{\tau}$ are sufficiently separated so that $Q=\abs{\sum_{j\neq i} \frac{(e_j^{\tau})^T\Delta_te_j^{\tau}}{\la_i^{\tau}-\la_j^{\tau}}}<1$ then $\abs{\mbf{a}_i-e_i^{\tau}}\leq C Q$, $\abs{\mbf{b}_i-e_i^{\tau}}\leq C Q$ and $\abs{\mbf{c}_i-\frac{1}{\sqrt{\tau}}(1,1,1,\dots,1)}\leq C Q$ for some bounded constant $C$. Since $\norm{\Delta_t}_2$ will typically be at most $O({\tau}-t)$, these estimates only control the large values $\la_i$ of the decomposition. 

Now specialize to the case when the ${Z}_t=A_t$, the (generalized) adjacency matrix of the time $t$ network $G_t$. Then there are typically many eigenvalues which are negative and small in magnitude and only a small percentage of large positive eigenvalues. Indeed, for a common ensemble of random networks, specifically those whose ${n}\times {n}$ adjacency matrix entries are Bernoulli random variables and equal to one with probability $p_{n}>>\frac1{{n}}$, such that $\sigma_{n}=\sqrt{{n} p_{n}(1-p_{n})}$ tends to infinity as ${n}$ does, the eigenvalues of $A_t$ have mean $-p_{n}$ and when mean centered and normalized to have unit standard deviation, their density limits almost surely as ${n}\to \infty$ to that of the semi-circle law whose density is $\begin{cases} \frac{1}{2\pi}\sqrt{4-x^2} & -2\leq x\leq 2\\ 0 & \text{otherwise}\end{cases}$ (\cite{DingEtAl:10}). In particular, the eigenvalues have mean $-p_{n}$, and only $o(\frac{1}{{n}})$ fraction of these are larger than $2 \sqrt{{n} p_{n}(1-p_{n})}-p_{n}$ with probability tending to $1$ as ${n}\to\infty$. Note that most networks in applications are quite sparse, or even bounded degree, which corresponds to the case that ${n} p_{n}$ grows very slowly, so that the effective upper bound of $2 \sqrt{{n} p_{n}(1-p_{n})}-p_{n}$ grows slowly as well. In almost every practical case, the percentage of eigenvalues larger than $o(\sqrt{{n}})$ decays to 0. 

We can therefore observe that by setting the embedding dimension ${d}$ to be a small fraction of ${n}$, we can still expect to capture most if not all of the large, eigenvalues of $A_t$ for a typical network. (Recall that typically $n<<r$.) The effective ratio $\frac{d}{n}$ for a given measure of prediction should generally be smaller for sparse networks seen in typical applications than for dense ones based on the number of large eigenvalues. As was explained in Subsections \ref{subsec:AdjacencyEmbed}, and \ref{subsec:GenAdj} for the path weighted adjacency matrices, it is these large eigenvalues of $A_t$ that capture their connectivity properties. As the previous perturbative analysis applies to these, we expect to recover these connectivity properties within layers in the embedding provided by our DynAOCPD algorithm, even for relatively small ${d}$ compared to ${n}$. However, there is also temporal information captured by the tensor modes as we will observe shortly. 

By contrast, when we choose ${Z}_t=L_t$, the Laplacian matrix of the time $t$ network $G_t$, the situation is more or less reversed. (Note the eigenvalues of $L_t$ are always nonnegative since it factors as $L_t=M_t^TM_t$ for the incidence matric $M_t$.) In the special case that $G_t$ is regular of degree $d$, the $i$-th largest eigenvalue $\la_i^t$ for $A_t$ corresponds to the $i$-th smallest eigenvalue for $L_t$ whose value is precisely $d-\la_i^t$. (In this case, $\la_i^t\leq d$ for all $i$.) The $\la_i^t$-eigenspace for $A_t$ is also precisely the $d-\la_i^t$ eigenspace for $L_t$. For nonregular networks, this relationship of the spectra of $A_t$ and $L_t$ still holds qualitatively. In particular, while there may be many small eigenvalues, typically this number will represent only a small fraction of ${n}$. We may quantify this more precisely using various results. First, Theorem 3.6 of \cite{XiaodongJiongsheng:01} shows the $k$-th largest eigenvalue of the Laplacian $L$ for a network is bounded below by,
\[
\lambda_{k} \geq \frac{1}{k}\left\{({n}-1)\left(2^{k} {n} \nu\right)^{\frac{1}{{n}-1}}-2 M\right\}
\]
where $M$ is the number of links and $\nu$ is the number of spanning trees of $G$. For many networks this is readily seen to be at least $O(\sqrt{{n}})$ when $k$ is $O({n})$. When $G_t$ is a random network as discussed above, the eigenvalues of $L_t$ have average ${n} p_{n}$ and when mean centered and normalized to have unit standard deviation, their density limits to that of a free convolution of the semi-circle law with a normal distribution (\cite{DingEtAl:10}). In particular these are clustered at the mean, and $o(\frac{1}{{n}})$ fraction of these are less than ${n} p_{n}-2\sigma_{n}$, which for typical choices of $p_{n}$ is $O(\sqrt{{n}})$, with probability tending to $1$ as ${n}\to\infty$. Proposition \ref{prop:resist} together with the discussion above it in Section \ref{sec:resist} shows that the collection of eigenvectors corresponding to small eigenvalues of $L_t$, i.e. large eigenvalues of $L_t^\dagger$ principally control the commute time between nodes of $G_t$. This same ordering arises in the optimization for the adjacency embedding in Section \ref{subsec:AdjacencyEmbed}. It follows that the relatively few eigenvectors for large eigenvalues of $A_t$, which correspond to those small eigenvalues of $L_t$, govern the connectivity properties of the network $G_t$. 

The perturbative analysis of the matrix case helps explain the meaning of the full tensor decomposition in our case. The largest prinicpal modes $\la_i$ of the tensor decomposition of $Z$ with $Z_t=A_t$ will correspond closely to the largest eigenvalues of $A_\tau$, at least when after applying our preconditioned linearly recurrent weighting in order to discount widely diverging data from the too distant past. Nevertheless these modes differ from simply reflecting the eigenvalues of the most temporally recent slice $Z_\tau=A_\tau$, as they are reinforced when there is a persistence over multiple slices. This is the point behind the contraction with the $\mathbf{c}_i$ vectors. The weights provide a means of controlling this process of this temporal connection capture for different applications. There may be more large valued $\la_i$ than there are for the eigenvalues of $A_\tau$ as some of the larger eigenvalues may split into multiple components that characterize distinct temporal relationships. This may help explain why in our experiments, the DynAOCPD embedding generally leads to improved task performance over the static embedding methods applied to $A_\tau$. 

The previous analysis can also be employed to explain why the analogous algorithm using $Z=L_t$, used in the DynL(O)CPD embeddings, instead of $Z_t=A_t$ must use the smallest modes $\la_i$. Principally, the eigenvectors of large eigenvalues of $L_t$ do not capture the connectivity between nodes but rather more subtle global structure of the network. The DynL(O)CPD embeddings are analogous to the resistance embedding which uses the reciprocals of the smallest eigenvalues. As explained in Subsection \ref{sec:resist}, one does not need to compute the inverse of a matrix to extract the $d$ smallest eigenvalues. However, for higher order tensors, the authors are not aware of any method to efficiently extract the $d$ smallest (in magnitude) modes of an (O)CP decomposition, unless perhaps it happens to be super-symmetric. Moreover, the above analysis strongly suggests that any generalized spectral components of $Z$ from which information about short-range network conectivity could be extracted would require finding these small modes. If this necessitates a full 
(O)CP decomposition of $Z$, then this quickly becomes prohibitively expensive as $r(Z)$ is typically significantly larger even than ${n}$. By contrast, there are known efficient techniques for extracting only the large modes $\la_i$ from a tensor decomposition via variations on least squares and Perron-Frobenius methods. Since the same network connectivity properties are equivalently captured by the spectrum of $A$ as of $L$, we will employ defer to using DynA(O)CPD embeddings in lieu of the DynL(O)CPD embeddings.

\subsection{Algorithm Implementations}

Below we give provide the implementation flow of the DynAOCPD algorithm. The DynACPD is the same with the obvious modification. The input stream of adjacency matrices $\set{A_t}$ may already include modifications such as the Katz generalized weighted adjacency matrices discussed in \ref{subsec:GenAdj}.

\begin{algorithm}
	\caption{Embedding algorithm}\label{euclid}
	\begin{algorithmic}[1]
		\Procedure{Embed}{$\set{A_t}_{t\in \mc{T}},{n},{\tau},d,W$}\Comment{$A_t\in \R^{{n}\times {n}}$, $W\in \R^\tau$, $\set{x_1,\dots,x_{n}}\subset\R^d$}
		\For{$t\in\mc{T}$}
			\State $Z_{\cdot,\cdot,t}\gets {A}_t$\Comment{$Z$ is ${n}\times {n}\times \tau$}
		\EndFor
		\State $(\Lambda,A,B,C)\gets OCPD(Z,d)$\Comment{$A,B$ are ${n}\times d$, $C$ is ${\tau}\times d$, $\Lambda$ is $d$-vector}
		\State $X\gets B.\op{diag}(\sqrt{\Lambda}*(C^T.W))$ \Comment{weighted convolution}
		\State \textbf{return } $X$\Comment{$X\in R^{{n}\times d}$ are embedding vectors}
		\EndProcedure
	\end{algorithmic}
\end{algorithm}

We implement the above algorithm in Python, and use the Tensorly library for the CPD algorithm. We also provide our own implementation of the OCPD algorithm based on a simple modification of the alternating least squares (ALS) algorithm, since at the time of writing we could not find an implementation in any of the common Python tensor analysis libraries.

Due to the large size of our datasets, we work with sparse matrices and operate on sparse tensors. While most of the libraries we use are equipped to handle sparse tensors, where standard libraries are not equipped to handle these operations we have written our own data type handlers and corresponding algorithms.

\section{Modeling of the Tasks}\label{sec:tasks}

Here we present some the details of the implementation of our three main classification/prediction tasks mentioned in the introduction. We assume we have already performed the embedding so that each node $i$ corresponds to a vector $v_i\in \R^d$. 

\subsection{Link Prediction}
Various authors have developed differing specific approaches to the link prediction task (see e.g. \cite{RichardEtAl:12,PachevWebb:17,KazemiPoole:18,ZhangChen:18,RossiEtAl:20}).
However, one widely used framework for link prediction applies a choice of classifier to a second choice of ``separation metric'' computed on all pairs of nodes in the embedding. Here the separation function need not be a metric distance, i.e. it  need not satisfy triangle inequality, but we usually insist that it be symmetric in its entries when we are considering undirected networks. 

There are two standard separation metrics that are often used in these problems which we term ``Hadamard'' and ``$L^2$''. The Hadamard separation metric is  simply the similarity matrix for the network embedding. Explicitly, we compute the Hadamard matrix,
\[
S=[S_{ij}]=[\inner{v_i,v_j}]=[\norm{v_i}\norm{v_j}\cos \theta_{ij}]
\]
where $v_i\in \R^d$ is the unit vector of the embedding of node $i$ and $\theta_{ij}\in [0,\pi]$ is the planar angle between the vectors $v_i$ and $v_j$. 

The $L^2$-separation metric is a true metric and is the usual Euclidean distance matrix with entries 
\[
D=[d(v_i,v_j)]=[\inner{v_i-v_j,v_i-v_j}^{\frac12}]=\sqrt{\norm{v_i}^2+\norm{v_j}^2 -2S_{ij}}
\]
In spite of the above simple relation between these two metrics, they often give different results depending on the data set (neither is always better than the other as we shall see in the experiments section).

Note that if the $v_i$ happen to all be unit vectors then the $L^2$-distance reduces to $d(v_i,v_j)=2\sin(\theta_{ij}/2)$. For small angle this metric is approximately $\theta_{ij}$ which is the standard symmetric (round) distance function on the unit sphere. (The similarity matrix of the distances themselves is sometimes referred to as the coarse curvature matrix of the finite configuration $\set{v_1,\dots,v_n}$, though not directly related to the differential geometric curvature tensor.)

The classifier will be a mapping $f:[0,\infty)\to [0,1]$ which can be viewed as an assignment of a probability $f(M_{ij})=w_{ij}\in [0,1]$ between all pairs of nodes based only on their separation metric $M$. We will not require $f$ to be monotone, and it may depend on the data set and choice of separation metric. 

For our choice of classifier, we use an $\ell_1$-regularized cross-validated logistic regression which is both trained and validated on a random sample of both existing and missing links with equal penalties for false positives as for false negatives. The training set, $E\subset \mc{V}\times \mc{V}$, consists of pairs of nodes with an equal number of links (i.e. the subset $E\cap \mc{E}_\tau$) and non-links ($E\cap \mc{E}_\tau'$). We create a training data set $\set{(x_e,y_e)}_{e\in E}$ where $x_e=\norm{v_{e_+}-v_{e_-}}$ for the embedding $v_i$ of node $i$ and $y_e\in\set{-1,1}$ is the indicator of whether $e$ is an link ($1$) or not a link ($-1$). The $\ell_1$ regularized loss function is 
\[
\min_{w,c} \norm{\vec{w}}_1+10\sum_{e\in E} \log(\exp(y_e(x_e w+c))+1).
\]
The above optimization is used on partitioned samples together with a cross-validation function consisting of the area under the receiver operating characteristic curve and using a standard sample partitioning scheme. This helps optimize the $w$ and $c$ parameters for a balanced combination of recall and precision. The reason for using the logistic regression is that the corresponding ramp function which more closely matches the separations seen in our separation metrics than a straight linear function would. Moreover it does not suffer from much of a time penalty compared to parameter estimation for more complicated nonlinear regressions, and it generally performs well in our experiments.

\subsection{Node Clustering}
The task of node clustering is the easiest to perform once the spacial embedding of the combined prediction network has been achieved. For this we run a spatial nearest neighbord algorithm with a given threshold to select cluster membershp of nodes. Recall that if we wish to cluster based on specific node labels or parameters then we encode proximity in the weights of the network which then leads to a different embedding. Hence we always work with the embedding which captures the spectral properties of the weighted network based on the weights which more naturally encode the network relations between nodes.  

As new nodes and links arrive which are assigned to a given cluster, then we must update the centroid accordingly. We will use the streaming k-means clustering algorithm in \cite{AilonEtAl:09} which uses parameters to control the time decay of importance attached to data in the calculations. Following \cite{YuEtAl:18}, we introduce a decay factor $\alpha$ which governs the decay of older data points in existing clusters when calculating the new cluster centers after absorbing new representation points(s) for the nodes. 

More specifically, we update the center assuming that there are $m_{0}$ points $\left\{v_{i}\right\}_{i=1}^{m_{0}}$ in an existing cluster and $m^{\prime}$ new points $\left\{v_{i}^{\prime}\right\}_{i=1}^{m^{\prime}}$ in $\mc{G}'$ to be absorbed by this cluster, the centroid $c$ can be updated in the following way
\[
c=\frac{\alpha c_{0} m_{0}+ \sum_{i=1}^{m^{\prime}} v_{i}^{\prime}}{\alpha m_{0}+ m^{\prime}}
\]
where $c_{0}$ is the previous cluster center. At $\alpha=1$ this would correspond to ordinary update of the centroid and at $\alpha=0$ only the new nodes would be used for the centroid. (Note the difference from the formula used in \cite{YuEtAl:18} where $\alpha=0.5$ corresponds to the usual centroid.) The decay parameter $\alpha$ is chosen as $0.5$ which modestly discounts the contribution from older nodes. (Effectively we are performing an exponentially decaying linear recurrence.)

\subsection{Anomaly Detection}
The anomaly detection problem in dynamic networks is an ancilliary problem to that of node clustering. It may be described as the following procedure. Given the node representations $v_i \in \mathbb{R}^{d}$ found from our embedding of derived from an initial temporal network $\mc{G}_0$, we first group these representative nodes into $k$ clusters. (For this we use the streaming k-means clustering algorithm described in the subsection on node clustering.) Given an updated temporal network $\mc{G}'$ with a set of newly included nodes and/or links with associated embedding representations $\set{v_i'}$ we can ask whether or not the newly included nodes naturally belong to one of the existing clusters based on a threshold for some distance function. If it does not, then the node will be labelled ``anomolous.'' 

For simplicity we measure the distance of a new node $v_i'$ to a fixed cluster $\set{v_{i_j}}_{j=1}^{m}$ by using the standard Euclidean ($\ell^2$) distance,  $\norm{\boldsymbol{c}-v_i'}_{2}$ where $c=\frac{1}{m}\sum_{j=1}^m v_{i_j}$  is the cluster center. The anomaly score for each point is reported as its closest distance to any of the cluster centers, and we choose a dataset specific threshhold on the distance for cluster inclusion.

\section{Experiments}\label{sec:experiments}

\subsection{Data sets}
\begin{table}[hb]
\begin{tabular}{|l|l|l|l|}
\hline
Datasets &  School&  Facebook& College  \\ \hline
Nodes $({n})$ &242  & 663&1899  \\ \hline
Edges &94---3736  &1688---3476  &358---12466  \\ \hline
Time Steps $(\tau=\abs{\mc{T}})$ &40  &9  &10  \\ \hline
\end{tabular}
\end{table}
For our experiment design, we use four real-world dataset, School (\cite{StehleEtAl:11}), College (\cite{PanzarasaEtAl:09}), Facebook (\cite{HajiramezanaliEtAl:19}). These are directly loaded into sparse tensor format which are then run through our algorithm and compared to baseline algorithms for the link prediction tasks.

\subsubsection{School}\footnote{networkrepository.com/dynamic.php}
This dataset was collected through radio-frequency identification devices to study the social patterns in primary school (\cite{StehleEtAl:11}). The devices recorded a contact between two people when they were within 1.5 meters proximity. The study included 232 students and 10 teachers for a total of 242 nodes. The devices recorded the activity between students and teachers over two consecutive days. From this data, we created 39 snapshot networks, each one having 242 nodes and between 94 and 3736 links and with consecutive snapshots equally spaced in time.

\subsubsection{Facebook}\footnote{github.com/VGraphRNN/VGRNN}
This Facebook wall posts was provided by authors of VGRNN and it was originally from \cite{HajiramezanaliEtAl:19}, with 9 time steps and 663 nodes after their data cleaning procedures. 

\subsubsection{College}\footnote{snap.standford.edu/data/CollegeMsg.html}
It is a dataset \cite{PanzarasaEtAl:09} of private messages from an online social network at UC, Irvine. The time span of the dataset is 193 days, with 1899 nodes. To ensure each snapshot will have an appropriate number of non-zero entries in its adjacency matrix, we created 10 snapshots from it, with the number of links ranging from 358 to 12466.

\subsection{Baseline Methods for Comparison}
In addition to our own proposed algorithms, the following baseline algorithms were applied for comparison.

\subsubsection{VGRNN} Variational Auto-encoders is a recently introduced algorithm, inspired by the Variational Graph Auto-Encoder (VGAE) algorithm (\cite{KipfWelling:16}) and the Variational Recurrent Neural Network (VRNN) algorithm (\cite{ChungEtAl:15}). It adopts the general structure of VRNN with each layer a VGAE unit. Lastly, to capture the temporal dependencies, it uses hidden states to compute the conditional priors for the generation process at each VGAE unit.

\subsubsection{DyanAERNN}
This algorithm (\cite{GoyalEtAl:20}) has an encoder-decoder structure. It consists of an encoder neural network comprised of a combination of both dense and ``Long Short Term Memory'' (LSTM) layers, and a fully connected network as a decoder. Instead of directly passing adjacency matrices for each snapshot, it first inputs them into a dense layer to create a lower dimensional hidden state, then passes these to the LSTM layers.

\subsubsection{``Res-Last'' - Resistance Embedding on last snapshot}
To provide evidence for the superior performance attributable to employing temporal information, we also compare our algorithm with three static embedding methods. The first one is the Laplacian Resistance Embedding described in Section \ref{sec:resist} using $[i]\mapsto (\frac{v_{ji}}{\sqrt{\la_j}})_j$ where $\la_j$ and $v_j$ are the eigenvalues and eigenvectors of the Laplacian matrix of the last snapshot.

\subsubsection{``Adj-Last'' - Spectral Adjacency Embedding on last snapshot}
The second static method is the Adjacency Embedding described in Section \ref{subsec:AdjacencyEmbed} using $[i]\mapsto (\mu_j v_{ji})_j$ where $\mu_j$ and $v_j$ are the eigenvalues and eigenvectors of the adjacency matrix of the last snapshot.

\subsubsection{``Res-Wt'' - Weighted Resistance Embedding}
We also consider the resistance embedding applied to the convolution of the Laplacian matrices of all the snapshots using a set of normalized Gaussian weights (with $\sigma=8$) applied to snapshots to create a single mollified snapshot representing a temporal average.

\subsubsection{``Adj-Wt'' - Weighted Adjacency Embedding}
We similarly apply the spectral adjacency embedding applied to the convolution of the adjacency matrices of all the snapshots using the same set of normalized Gaussian weights ($\sigma=8$).

\subsubsection{GCN}
The third static embedding method is the Deep Graph Infomax (DGI) algorithm (\cite{VelickovicEtAl:18}), which is an unsupervised learning method, together with a Graph Convolutional Network (GCN) dimension reduction layer, relying on maximizing the mutual information. Since the DGI algorithm is simply used to make the GCN unsupervised we will call this by the more familiar GCN term. 

\subsubsection{DynACPD+GCN}
Also to compare with our proposed method, we use DGI to first reduce the dimension of each snapshots' adjacency matrix and with these we build the tensor for DynACPD instead of using the original adjacency matrix. The remaining steps of the DynACPD algorithm are then applied to get the final embedding as usual. 

\subsection{Parameter Settings}
Through the experiments, we search for the best performance dimension of our algorithms and the Laplacian Resistance embedding in the set $d\in\set{8,16,32,64,128}$. The optimum consistently holds at dimension $d=128$.  For convolution weights we use a Gaussian with  standard deviation parameter $\sigma$ set to same value as for the weights used in the linear recursion pre-processing. We optimize our algorithms over $\sigma\in\set{8,16,32}$ as well as over the two binary parameters of nonnegative vs standard CPD algorithms and unit normalized vs. unnormalized embeddings. 

For the VGRNN algorithm, we set its two GCN layers to fixed sizes $(32,16)$ and training epochs to $1500$ with learning rate equal to $0.01$ across all data-sets. For GCN and DynACPD+GCN, the DGI's two GCN layers are also set to $(32,16)$, with other setting exactly the same as for VGRNN. (These dimension values are the respective defaults for the algorithms as determined in their original papers which were found to be a well tuned balance between performance and time cost.)

\begin{figure}[H]
\subfloat[fig 1]{\includegraphics[width = 3in]{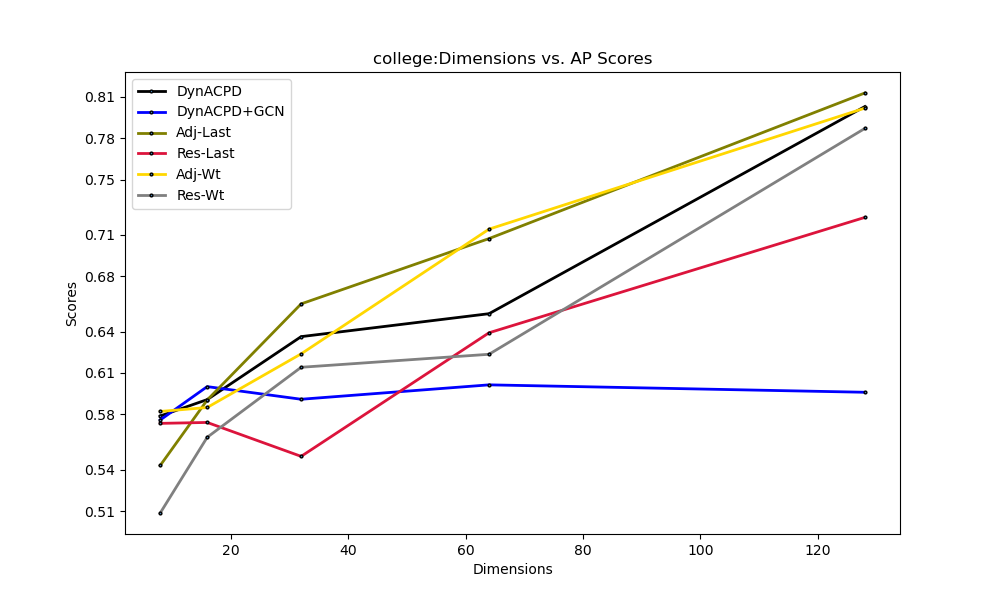}} 
\subfloat[fig 2]{\includegraphics[width = 3in]{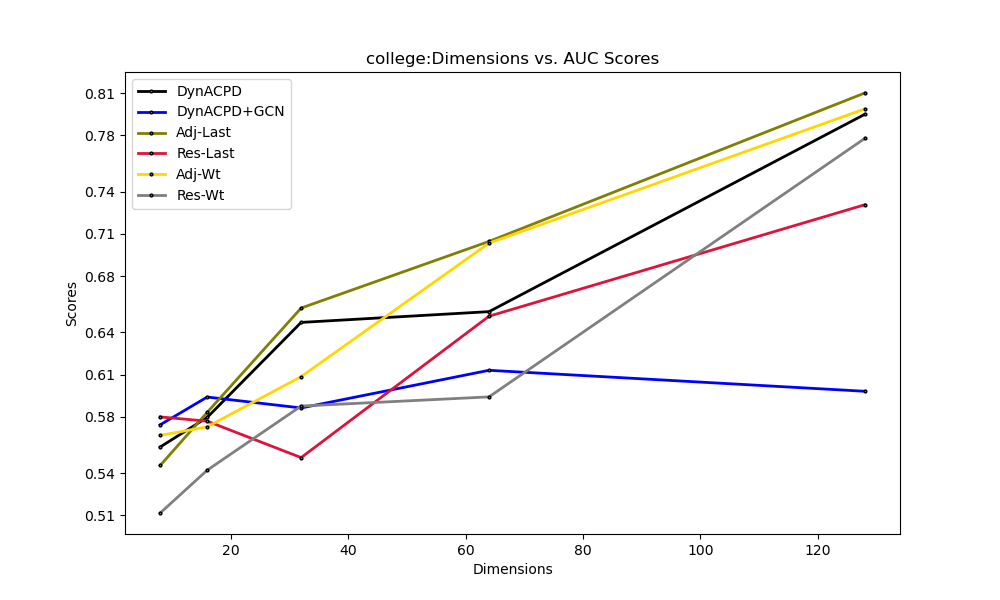}}\\
\subfloat[fig 3]{\includegraphics[width = 3in]{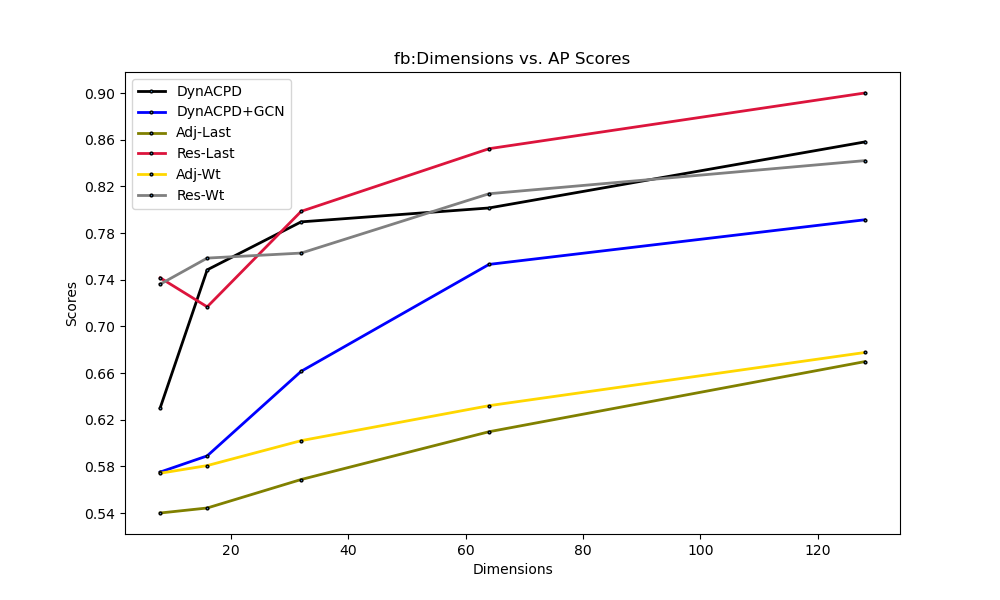}}
\subfloat[fig 4]{\includegraphics[width = 3in]{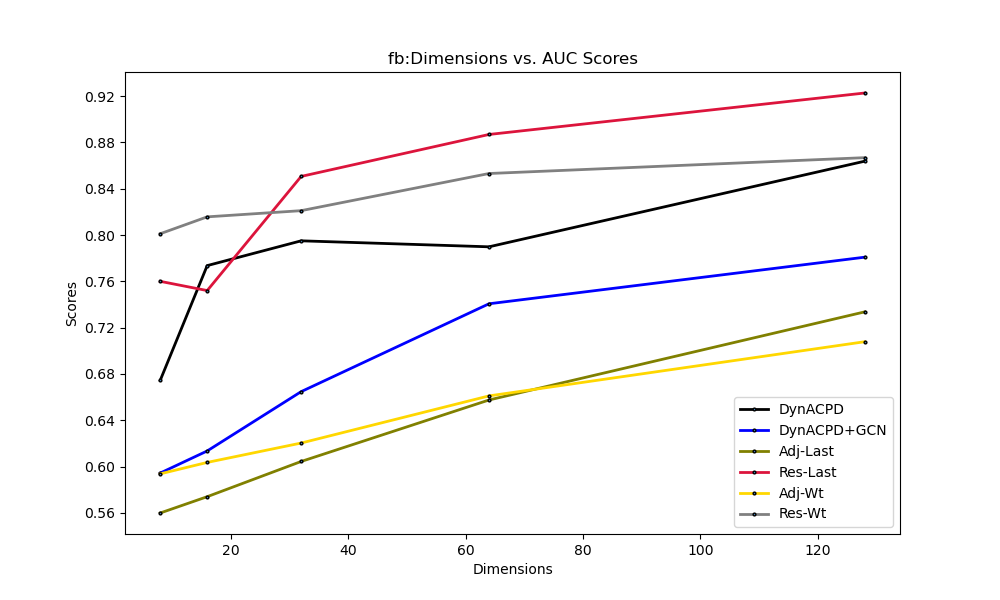}} \\
\subfloat[fig 5]{\includegraphics[width = 3in]{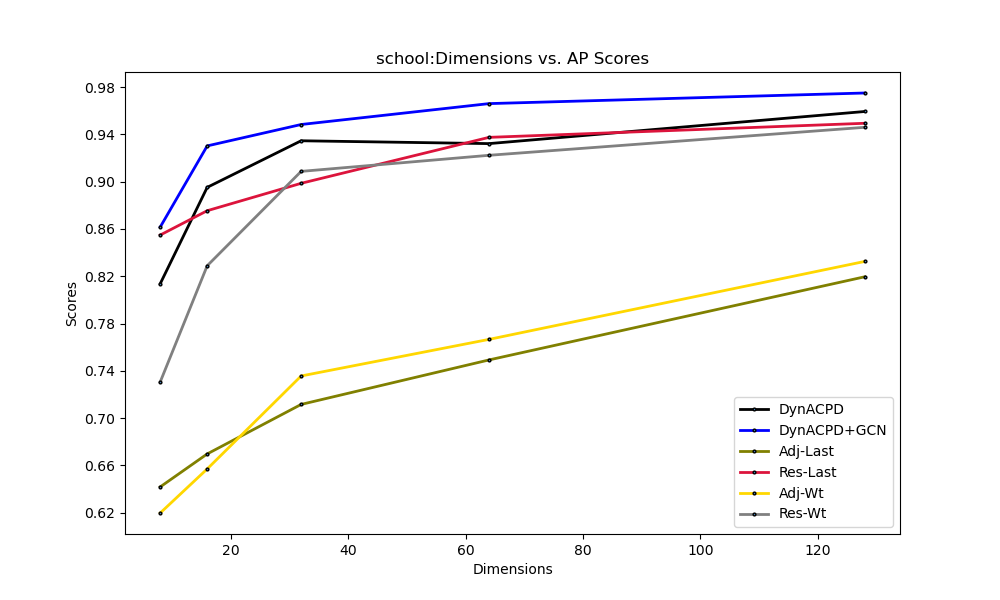}}
\subfloat[fig 6]{\includegraphics[width = 3in]{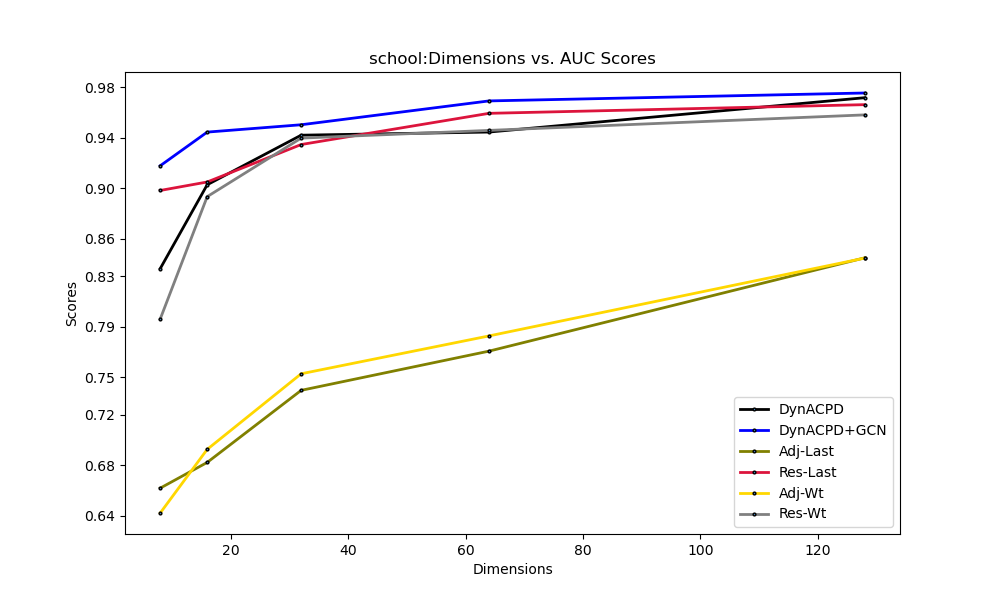}} 
\caption{Dimensions vs. AP/AUC scores for $L^2$ norm.}
\label{fig:L2}
\end{figure}

\begin{figure}[H]
\subfloat[fig 1]{\includegraphics[width = 3in]{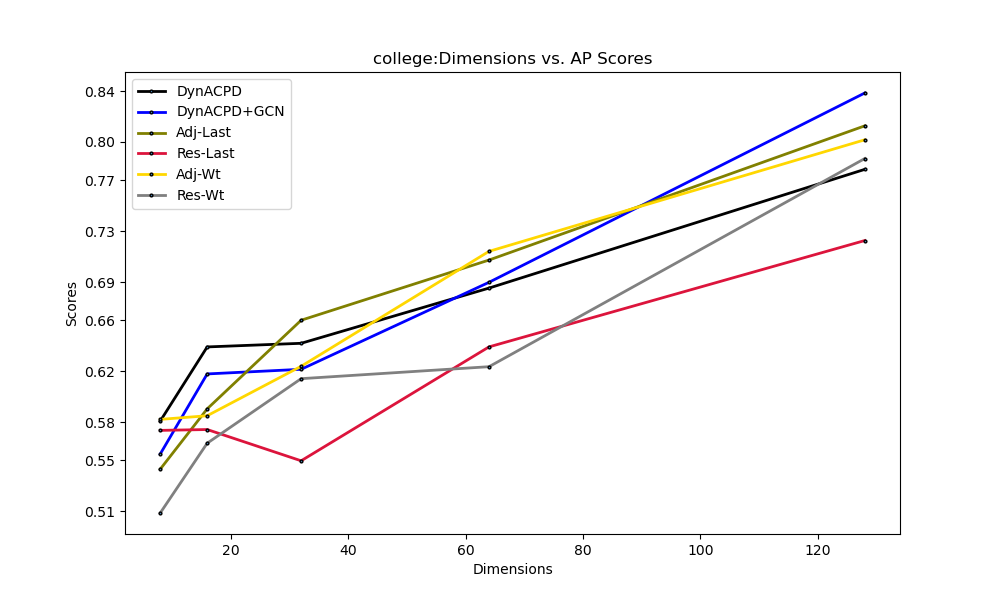}} 
\subfloat[fig 2]{\includegraphics[width = 3in]{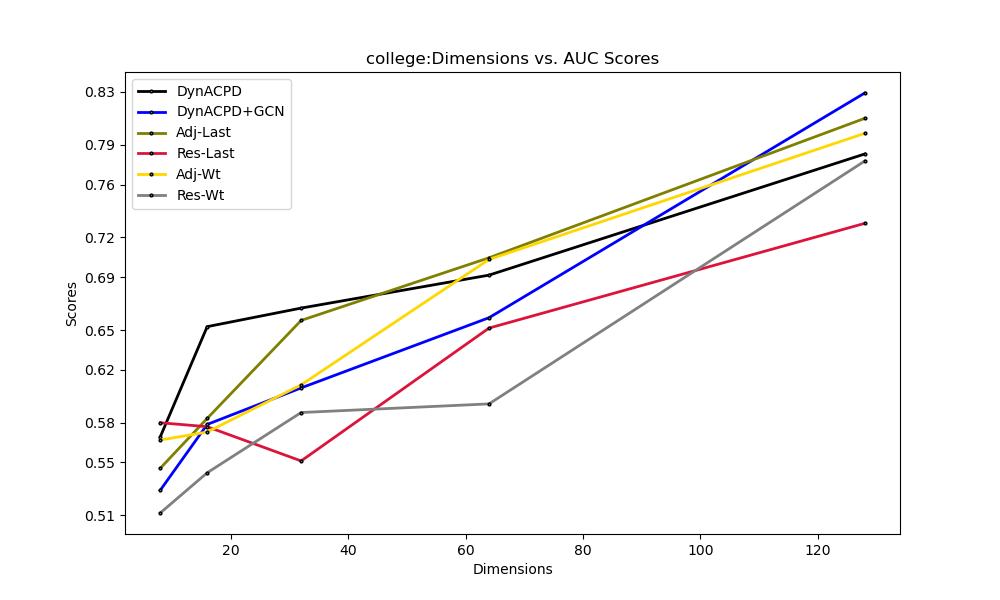}}\\
\subfloat[fig 3]{\includegraphics[width = 3in]{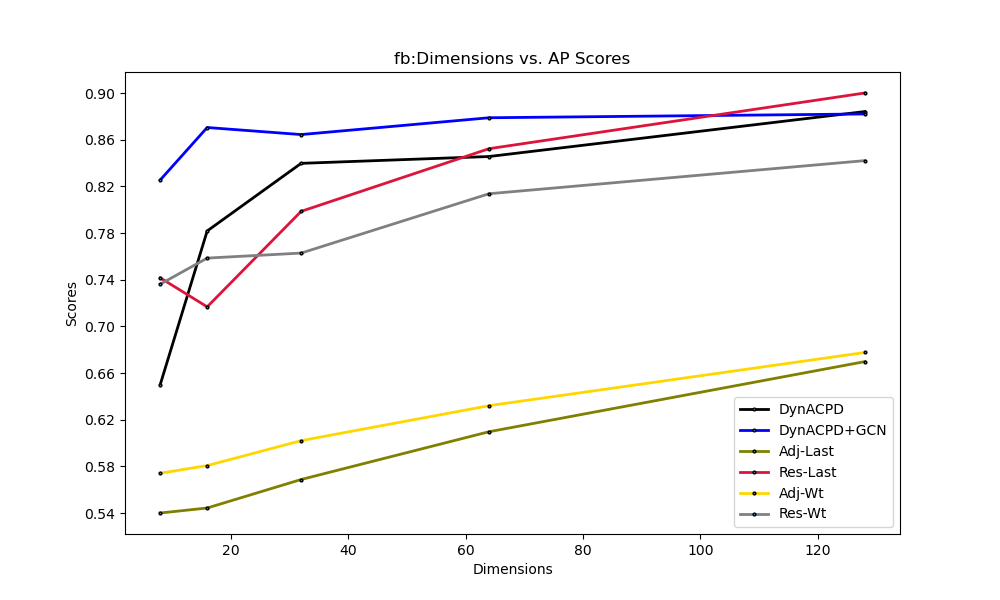}}
\subfloat[fig 4]{\includegraphics[width = 3in]{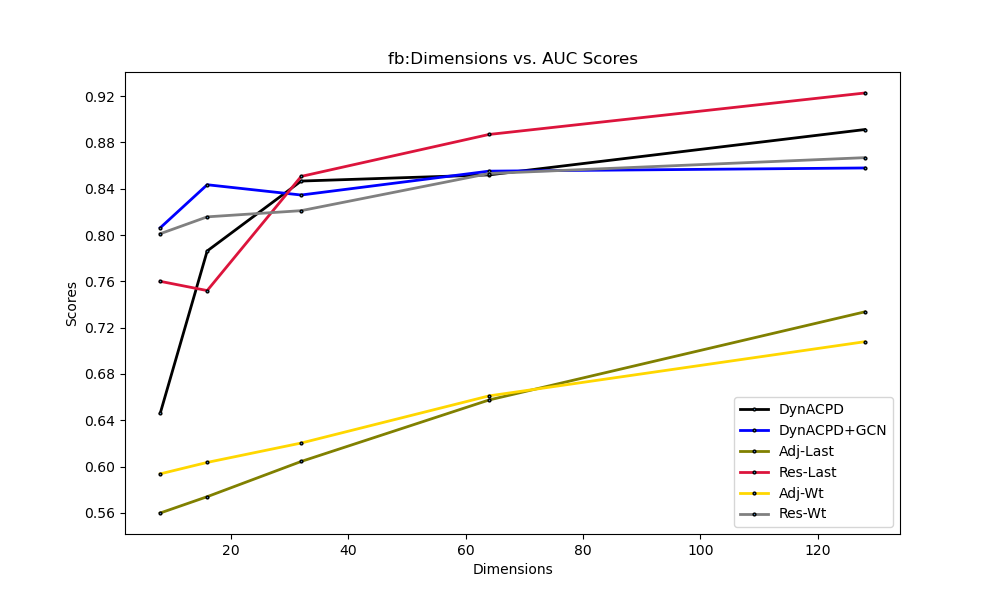}} \\
\subfloat[fig 5]{\includegraphics[width = 3in]{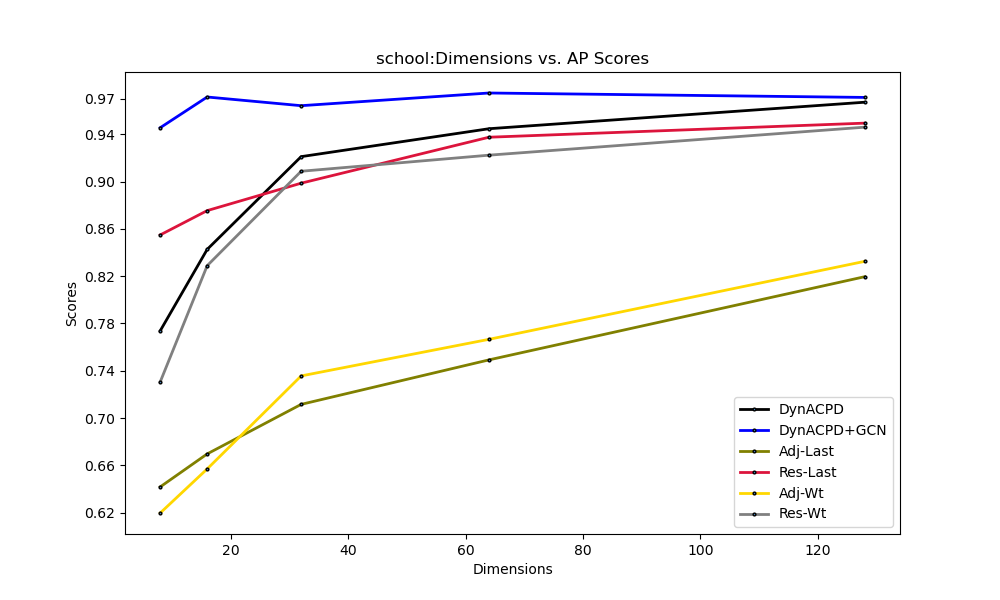}}
\subfloat[fig 6]{\includegraphics[width = 3in]{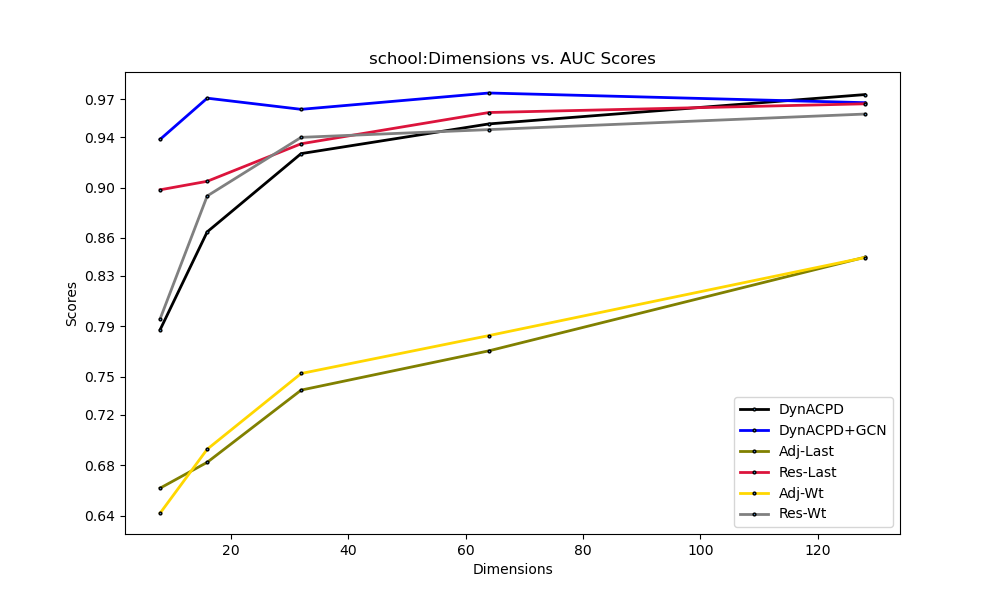}}
\caption{Dimensions vs.  AP/AUC scores for Hadamard norm.}
\label{fig:Hadamard}
\end{figure}

In Table \ref{tab:compL2} and Table \ref{tab:compHad} we report the scores for the link prediction task with the dimension parameter set to 192 across all algorithms (except for VGRNN which does not have the same dimension parameter) for the $L^2$ Hadamard separation metrics respectively. This dimension setting leads to a comparable running time between the DynA(O)CPD algorithms and VGRNN (see Figure \ref{fig:RunTime}).  A full scores trend network over all searched dimensions for this task using the $L^2$ and Hadamard norms, respectively, for separation in the link prediction task is shown in Figures \ref{fig:L2} and \ref{fig:Hadamard}, respectively. (Again VGRNN is not compared in these figures as its layer dimension parameters are optimized independently.)

\subsection{Comparison}

In this section, we present the experimental results from the link prediction task. These scores provide some justification for the intuition of our methods to integrate the local topology and temporal network information into a unified network embedding. In Tables \ref{tab:compL2} and \ref{tab:compHad}, we highlight the top three scores, in red blue and green respectively, and in Tables \ref{tab:equalL2} and \ref{tab:equalHad}, we highlight the top two scores in red and blue respectively. The Python code is made available at \url{github.com/yw109iu/DynCPD}.

We report the scores with both Hadamard and $L^2$ binary operators, since they represent two very different approaches to how points in the embedding are spearated. (Indeed, each is a canonical representative of a family of common norms that arise naturally from practical examples.) Interestingly, each of these two norms provides preferable separation over the other depending on the dataset and method. 

\begin{table}[H]
\smaller\smaller
\centering
\caption{L2 Scores, $d=32$}
\begin{tabular}{ |c|c|c|c|c|c|c| } 
\hline
Metrics & Methods & School & Facebook & College \\
\hline
\multirow{3}{4em}{AP}  
& \bf{DynAOCPD} & 0.9411 & 0.7965 & \cB{0.6363}  \\
& \bf{DynACPD} & 0.9454 & \cR{0.8107} & 0.6289 \\
& Res-Wt& 0.9087 & 0.7629 & 0.6141  \\
& Adj-Wt& 0.7356 & 0.6021 & 0.6239  \\
& GCN-Last (DGI) & 0.7929 & 0.7153 & 0.5652  \\
& GCN-Wt (DGI) & 0.7454 & 0.7201 & 0.6134  \\
& \bf{DynAOCPD+GCN} & \cB{0.9609} & 0.6957 &0.5841 \\
& \bf{DynACPD+GCN} & \cR{0.9628} & 0.6531 &0.5847 \\
& Adj-Last & 0.7394 & 0.5688 & \cR{0.6600} \\
& Res-Last & 0.8986 & \cB{0.8020} & 0.5497 \\
\hline
\multirow{3}{4em}{AUC}  
& \bf{DynAOCPD} & 0.9476& 0.8051 & \cB{0.6471}  \\
& \bf{DynACPD} & 0.9454 & 0.8119 & 0.6392  \\
& Res-Wt& 0.9397 & \cB{0.8211} & 0.5876  \\
& Adj-Wt& 0.7526 & 0.6203 & 0.6085  \\
& GCN-Last (DGI) & 0.8702 & 0.7574 & 0.5548  \\
& GCN-Wt (DGI) & 0.8324 & 0.7845 & 0.6102  \\
& \bf{DynAOCPD+GCN} & \cB{0.9632} & 0.6844 &0.5752 \\
& \bf{DynACPD+GCN} & \cR{0.9655} & 0.6495 &0.5747 \\
& Adj-Last & 0.7394 & 0.6044 & \cR{0.6573} \\
& Res-Last & 0.9347 & \cR{0.8525} & 0.5510 \\
\hline 
\end{tabular}
\label{tab:compL2}
\end{table}

\begin{table}[H]
\smaller\smaller
\centering
\caption{Hadamard Scores, $d=32$}
\begin{tabular}{ |c|c|c|c|c|c|c| } 
\hline
Metrics & Methods & School & Facebook & College \\
\hline 
\multirow{3}{4em}{AP}  
& \bf{DynAOCPD} &0.9211& 0.8391 & 0.6485  \\
& \bf{DynACPD} & 0.9218& 0.8481 & \cR{0.6742} \\
& Res-Wt& 0.9318 & 0.7886 & 0.5852 \\
& Adj-Wt& 0.7694& 0.8103 & \cB{0.6594}  \\
& GCN-Last (DGI) & 0.7551& 0.7452 & 0.5481  \\
& GCN-Wt (DGI)& 0.6628 & 0.76015 & 0.5507  \\
& \bf{DynACPD+GCN} & \cR{0.9723} & \cB{0.8786} &0.6535 \\
& \bf{DynAOCPD+GCN} & \cB{0.9719} & \cR{0.8818} &0.6291 \\
& Adj-Last & 0.6915 & 0.8878& 0.6274 \\
& Res-Last & 0.9405 & 0.6750 & 0.5956\\
\hline 
\multirow{3}{4em}{AUC}  
& \bf{DynAOCPD} &  0.9269& 0.8430 & 0.6700  \\
& \bf{DynACPD} & 0.9301 & 0.8495 & \cR{0.6987}  \\
& Res-Wt& 0.9488 & 0.8151 & 0.5665  \\
& Adj-Wt& 0.7201 & 0.7568 & \cB{0.6403}  \\
& GCN-Last (DGI) & 0.8070 & 0.7736 & 0.5403  \\
& GCN-Wt (DGI) & 0.7212 & 0.7947 & 0.5481  \\
& \bf{DynACPD+GCN} & \cR{0.9724} & \cB{0.8598} &0.6268 \\
& \bf{DynAOCPD+GCN} & \cB{0.9710} & \cR{0.8635} & 0.5980 \\
& Adj-Last & 0.6453 & 0.8419 & 0.6252 \\
& Res-Last & 0.9557 & 0.7619 & 0.6043 \\
\hline
\end{tabular}
\label{tab:compHad}
\end{table}

We note some highlights from table \ref{tab:compL2}, comprising the scores with respect to the $L^2$ operator. For the school dataset, VGRNN has the best AUC socres and the second best AP scores, while our DynACPD+GCN claims the best AP scores. Generally speaking, in terms of AP and AUC scores, the difference between our proposed  methods and VGRNN is fairly small. Also, the Resistance Embedding on the last slice (Res-Last) is not far behind them. It also becomes clear that the other baselines generally perform much worse.  

For the Facebook data set, the situation is different, although VGRNN still claims the best scores in both AP and AUC, Res-Last is in the second place, with our proposed DynACPD methods very closely behind. However, the DynACPD+GCN method performs much worse. Since DynACPD works well as expected, one possible explanation for this difference on the Facebook dataset is that the higher link density of this dataset causes the pre-compression DGI layer to significantly alter the structure of the largest eigenvalues of the resulting adjacency matrix of the corresponding virtual network. In other words, the DGI compression for its given  optimized GCN parameters (using an entropy loss function) may destroy a significant amount of the topology, especially short cycles, in the snapshot networks. 

For the College data set, VGRNN still has the best scores, but instead of Res-Last, it is the Adjacency Embedding on the last slice (Adj-Last) that claims the second place though with our DynACPD method very close behind. Again, the pre-GCN encoded DynA(O)CPD methods lag behind significantly.  

Note that if the temporal network has the approximate Markov property, in that a future time slice only depends almost exclusively on the preceding slice and is otherwise nearly independent (with probability nearly 1) of the older time slices, then we would expect Res-Last or Adj-Last to perform as well as or even better than dynamic algorithms. We expect this property would likely account for most instances where these static algorithms are nearly top performers for a given dataset. As there is a cyclical pattern in the school dataset it most strongly fails to have the Markov property among the three datasets and as expected, the static methods perform poorly on it.

Many of the trends from Table \ref{tab:compHad}, comprising the scores with respect to the Hadamard operator are similar to those in Table \ref{tab:compHad}. However, interestingly, there is a pronounced improvement in the scores of DynACPD+GCN method when using the Hadamard norm to separate points. Also, the Res+Last scores drop significantly on the Facebook dataset. This may be explained by the close connection that the Resistance Embedding has to the $L^2$-norm for point separation by Proposition \ref{prop:resist}.

\begin{figure}[H]
\subfloat[fig 1]{\includegraphics[width = 3in]{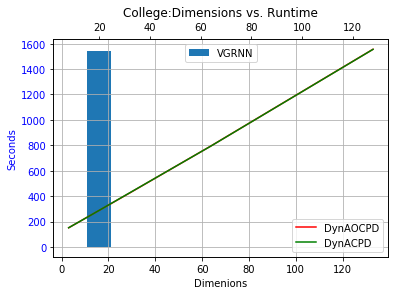}} 
\subfloat[fig 2]{\includegraphics[width = 3in]{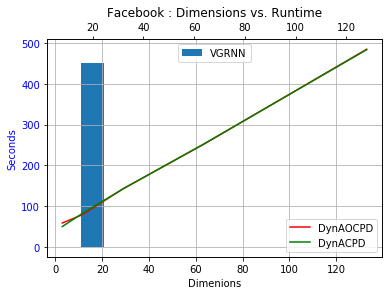}}\\
\subfloat[fig 3]{\includegraphics[width = 3in]{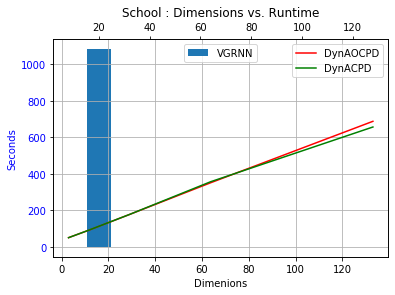}}
\caption{Running Time}
\label{fig:RunTime}
\end{figure}

As shown, the VGRNN algorithm performs exceptionally well for link prediction. However, there is a downside in that both the VGRNN algorithm and the DynAERNN algorithm take significantly longer to converge than any of the other algorithms considered unless the embedding dimension parameter becomes large. In Figure \ref{fig:RunTime}, we can explicitly see the differences between their running time and those of DynACPD and DynAOCPD as the dimension parameter is increased for the latter two. (DynACPD and DynAOCPD have nearly identical running times.)

If we change the dimension parameter to 128, then the DynA(O)CPD algorithms still run faster. However, their scores as shown in Tables \ref{tab:equalL2} and \ref{tab:equalHad} for the $L^2$ and Hadamard norms respectively, become comparable and sometimes exceed those of the VGRNN algorithm.

\begin{table}[htbp]
\smaller\smaller
\centering
\caption{L2 Scores, $d=128$ (roughly equal running time)}
\begin{tabular}{ |c|c|c|c|c|c|c| } 
\hline
Metrics & Methods & School & Facebook & College \\
\hline
\multirow{3}{4em}{AP}  
&  DynAERNN & 0.5817 & 0.6144 & 0.5578  \\
&  VGRNN & {0.9623} & \cB{0.8845} &   {0.7248}\\ 
& \bf{DynAOCPD} &\cB{0.9880}& 0.8183 &  {0.7984}\\
& \bf{DynACPD} & \cR{0.9882}& {0.8339} & 0.7111  \\
& Res-Wt& 0.9460 & 0.8421 & 0.787 \\
& Adj-Wt& 0.8326 & 0.6777 & \cB{0.8017} \\
& Adj-Last & 0.8196 & 0.6698 & \cR{0.8127} \\
& Res-Last & 0.9494 & \cR{0.9001} & 0.7226 \\
\hline 
\multirow{3}{4em}{AUC}   
&  DynAERNN  & 0.5723 & 0.6326 & 0.5453  \\
& VGRNN & {0.9689} & \cB{0.8842} &   {0.7545} \\ 
& \bf{DynAOCPD} &  \cR{0.9888} & 0.8353 & \cR{0.8138} \\
& \bf{DynACPD} & \cB{0.9882} & {0.8496} & 0.7093 \\
& Res-Wt& 0.9582 & 0.8669 & 0.7778 \\
& Adj-Wt& 0.8445 & 0.7078 & {0.7987} \\
& Adj-Last &0.8447 & 0.7336 & \cB{0.8101} \\
& Res-Last & 0.9664 & \cR{0.9227} & 0.7306 \\
\hline
\end{tabular}
\label{tab:equalL2}
\end{table}

\begin{table}[htbp]
\smaller\smaller
\centering
\caption{Hadamard Scores, $d=128$ (roughly equal running time)}
\begin{tabular}{ |c|c|c|c|c|c|c| } 
\hline
Metrics & Methods & School & Facebook & College \\
\hline
\multirow{3}{4em}{AP}  
&  DynAERNN  & 0.5842& 0.7107 & 0.5728  \\
& VGRNN & {0.9788} & \cR{0.9560} & 0.7659  \\ 
& \bf{DynAOCPD} &\cB{0.9851}& 0.9092 & \cR{0.8410}  \\
& \bf{DynACPD} & \cR{0.9873} & {0.9098} & \cB{0.8280}  \\
& Res-Wt& 0.9597 & 0.9186 & 0.7864 \\
& Adj-Wt& 0.7995 & 0.8460 & 0.6793 \\
& Adj-Last & 0.7388 & 0.9071 & {0.8002} \\
& Res-Last & 0.9663 & \cB{0.9348} & 0.5737 \\
\hline 
\multirow{3}{4em}{AUC}   
&  DynAERNN  & 0.5801 & 0.6943 & 0.5445  \\
& VGRNN & {0.9821} & \cR{0.9597} & {0.7942}  \\ 
& \bf{DynAOCPD} &  \cB{0.9843} & 0.8901 & \cR{0.8340}  \\
& \bf{DynACPD} & \cR{0.9862} & {0.8913} & \cB{0.8214}  \\
& Res-Wt& 0.9660 & 0.9164 & 0.7767 \\
& Adj-Wt& 0.7645 & 0.7849 & 0.6563 \\
& Adj-Last & 0.7074 & 0.8670 & 0.7858 \\
& Res-Last & {0.9721} & \cB{0.9430} & 0.5752 \\
\hline
\end{tabular}
\label{tab:equalHad}
\end{table}

\section{Related Work}
\subsection{Static Graph Representation Approaches}
Node embeddings in static networks have been widely explored since the famous DeepWalk\cite{PerozziEtAl:14} algorithm first appeared. This builds on the idea of the skip-gram model\cite{MikolovEtAl:13}, first introduced in the domain of natural language processing, which aims to learn vector representations for words. Deepwalk and Node2vec both extend this idea by considering the nodes as the 'words', the network as the 'document' and add a random-walk to sample ordered node sequences as 'sentences'.  Much earlier than this, Spectral methods had been introduced for the purpose of dimension reduction\cite{BelkinNiyogi:03} and clustering problems\cite{NgEtAl:02}. Recently, matrix factorization methods attempt to formulate different proximity matrices based on the n-hop transitional probability matrix\cite{ZhangEtAl:18}, Katz Index\cite{OuEtAl:16}, Personalized Pagerank\cite{OuEtAl:16}, Common Neighbors\cite{OuEtAl:16} and Adamic-Adar\cite{OuEtAl:16} methods. Inspired by the success of neural networks in processing grids of images, various network-based neural networks have been introduced to learn node embeddings with convolutional networks\cite{KipfWelling:17}, attention  networks\cite{VelickovicEtAl:18}, and variational auto-encoders\cite{KipfWelling:16}.

Also node embedding algorithms in attributed networks\cite{BandyopadhyayEtAl:19} and heterogeneous networks\cite{HusseinEtAl:18} have been explored widely. Despite their success, many real-world scenarios are essentially dynamic, for example, relationships in a social network\cite{SarkarMoore:06},  spatio-temporal traffic prediction\cite{ZhengEtAl:20} and progressions of aging related genes \cite{LiMilenkovic:20}. Algorithms for static networks generally fail to consider the evolution of network structures, and thus lack the ability to capture any time-dependent information which can affect the performance of downstream tasks. 

\subsection{Dynamic Graph Representation Approaches}
Recently, dynamic node embedding algorithms have become another highly active research area \cite{HajiramezanaliEtAl:19}\cite{KumarEtAl:19}. Most of these newly introduced algorithms represent the dynamic networks as a series of static snapshots, where each snapshot represent the network in a discrete-time interval. The key problem of such algorithm is how to model the time dependencies, different ideas has been proposed to solve it. In an attempt to extend the basic idea of Deep-Walk into the dynamic environment, several versions of 'temporal random walks' have been introduced to sample ordered node sequences across time-points while balancing local topology\cite{NguyenEtAl:18}.  Some other algorithms capture time-dependencies by modeling the formation of temporal neighborhoods. Both the Hawkes Process\cite{ZuoEtAl:18} and the Triadic Closure Process \cite{ZhouEtAl:18} fall into this category. After proving its success in other time-series processing problems, some Recurrent Neural Network based algorithms were introduced recently to tackle the dynamic node embedding problem. One of them is the Variational Graph Recurrent Neural Networks\cite{HajiramezanaliEtAl:19}. Introduced to better capture temporal-dependencies, it utilized the last time points hidden layer as a prior distribution for the variational network auto-encoder's generation process at each time point. Despite the fact that it has good performance scores in downstream tasks, training an RNN based model can be both time and computation consuming.

\subsection{Laplacian for Tensors as Multigraphs}

Another approach which interpolates the static network approach and the dynamic network approach is to incorporate the time slices into a single large network which also incorporates the intra-network connections. If we write the coupling network as $G_C$  and the time-slice networks as $G_t$ then the adjacency matrix for the entire network $G=G_C\coprod_{t\in \mc{T}}G_t$ is
\[
A_G=A_C+\oplus_{t\in \mc{T}} A_t.
\]
Note that if we label the nodes of the slices in order to match as closely as possible then $A_C$ is $n\times n$-block 3-banded with $0$ diagonal blocks and nearly diagonal off-diagonal blocks.
The resulting Laplacian matrix,
\[
L_G=D_C-A_C+\oplus_{t\in \mc{T}} D_t-A_t=L_C+\oplus_{t\in \mc{T}}L_t,
\]
therefore consists of nearly diagonal off-diagonal $n\times n$ blocks with ${\tau}$ diagonal $L_t$ blocks. Thus the sparseness is dominated by the sparseness of the $L_t$. Some sources, especially those in connection with multiplex networks, call $L_G$ the ``supra-Laplacian'' (see e.g. \cite{SoleRibaltaEtAl:13,KunchevaMontana:17,GomezEtAl:13,CozzoMoreno:16}) to emphasize the network partition structure. However, we will consider $L_G$ as simply the Laplacian of the full network $G$. 

The normalized laplacian for the total network in the weighted but undirected case has the form
\[
\mc{L}_G=\begin{pmatrix}
I - D_1^{-\frac12}A_1 D_1^{-\frac12} & -D_1^{-\frac12}W_{12}D_2^{-\frac12} & &0 \\
-D_2^{-\frac12}W_{21}D_1^{-\frac12} & I-D_2^{-\frac12}A_2 D_2^{-\frac12} & -D_2^{-\frac12}W_{23}D_3^{-\frac12} &  \\
&&& \\
   \ddots & \ddots &\ddots & \ddots\\
   &&& \\
0 & &-D_{\tau}^{-\frac12}W_{{\tau},{\tau}-1}D_{{\tau}-1}^{-\frac12} & I-D_2^{-\frac12}A_2 D_2^{-\frac12} 
\end{pmatrix}
\]
Where $I$ is the ${n}\times {n}$ identity matrix, and $D_t$ is the diagonal matrix of degrees of the slice network $G_t$ and $W_{t,s}$ is the matrix of weights on links from the slice $G_t$ to the adjacent slice $G_s$ where $s=t+1$ or $s=t-1$. 

Note that in our case each node in the $t$ slice connects exactly only to the corresponding node in the $t+1$ slice. Hence $W_{t,t+1}=W_{t+1,t}$ is always diagonal and $-D_t^{-\frac12}W_{t,t-1}D_{t-1}^{-\frac12}=W_{t,t-1}D_{t-1}^{-1}$ with diagonal entries $\frac{w_{ii^+}}{\sum_k w_{ik}}$ where the index $i$ and $i^+$ are understood to be the corresponding indices in the $G_t$ and $G_{t+1}$ subnetworks.

The corresponding normalized Laplacian of $G$ in the directed case has similar structure,
\[
\mc{L}_G=\begin{pmatrix}
\mc{L}_1 & - \hat{W}_{1,2} & & & \\
- \hat{W}_{2,1} & \mc{L}_2 & - \hat{W}_{2,3} & & \\
&  \ddots & \ddots &\ddots &\\
& & &- \hat{W}_{{\tau},{\tau}-1} & \mc{L}_{\tau} 
\end{pmatrix}
\]
with diagonal block entries being $\mc{L}_t$, the normslized directed Laplacian of the slices and $\hat{W}_{t,s}=\hat{W}_{s,t}$ are diagonal with entries $\frac{\frac12 (w_{ii^+}+w_{i^+i})}{\sqrt{\sum_k w_{i k}}\sqrt{\sum_k w_{i^+ k}}}$.

Unfortunately, the dimension ${n}\tau \times {n}\tau$ for $\mc{L}_G$ makes calculations unwieldy, even when using sparse algorithms. However, a more fundamental problem for spectral methods is that the spectrum of $\mc{L}_G$ will nearly be $\tau$ duplicates of the spectrum of one ${n}\times {n}$ block, and so the spectral information must be manually combined in order to extract useful information on how to best embed for prediction.

\section{Conclusion}

We have provided an algorithmic framework for incorporating temporal
dependencies into network embeddings for the purposes of prediction tasks. The framework applies tensor decomposition methods to discrete time-dependent networks to extract analogs of spectral data that capture link connectivity features for dynamically varying networks. 

This generalized spectral data provides the basis for a single network embedding analogous to the resistivity embedding for static networks which appropriately incorporates temporal changes within the discrete-time
dynamic network. We demonstrate the effectiveness of our dynamic network embedding algorithms for temporal link prediction in multiple real-world networks.

Overall, the proposed method achieves an average gain of 7.6\%
across all baseline methods and dynamic network datasets. Our results indicate that generalized spectral tensor methods can be effectively used to improve the campture of link relationships in dynamic networks for the purpose of improving link prediction tasks. In future work we hope to investigate fully weighted time dependencies between nodes of time-slice networks in dynamic network.

\providecommand{\bysame}{\leavevmode\hbox to3em{\hrulefill}\thinspace}

\providecommand{\bysame}{\leavevmode\hbox to3em{\hrulefill}\thinspace}
\providecommand{\MR}{\relax\ifhmode\unskip\space\fi MR }
\providecommand{\MRhref}[2]{%
	\href{http://www.ams.org/mathscinet-getitem?mr=#1}{#2}
}
\providecommand{\href}[2]{#2}

%\bibliographystyle{alpha}
%\bibliography{TensorDecomp}
\newcommand{\etalchar}[1]{$^{#1}$}

\end{document}